\documentclass[10pt]{article}
\newcommand{\documentdate}{27 VIII 2026}
\usepackage{a4wide,latexsym,graphicx,amsmath,amssymb,varioref}
\usepackage{graphics,xcolor,dsfont}

\DeclareMathAlphabet{\pazocal}{OMS}{zplm}{m}{n}
\newcommand{\calA}{{\pazocal{A}}}

\newcommand{\calF}{{\pazocal{F}}}

\newcommand{\calM}{{\pazocal{M}}} 
\newcommand{\calN}{{\pazocal{N}}} 
\newcommand{\calO}{{\pazocal{O}}}

\newcommand{\calS}{{\pazocal{S}}} 
\newcommand{\calU}{{\pazocal{U}}}

\newcommand{\beqn}[1]{\begin{equation}\label{#1}}
\newcommand{\eeqn}{\end{equation}}
\newcommand{\req}[1]{(\ref{#1})}
\newcommand{\ms}{\;\;\;\;}
\newcommand{\tim}[1]{\;\; \mbox{#1} \;\;}

\newtheorem{theorem}{Theorem}[section]
\newtheorem{assumption}{Assumption}
\newtheorem{lemma}[theorem]{Lemma}

\newcommand{\numsection}[1]{\section{#1}\setcounter{equation}{0}}
\newtheorem{corollary}[theorem]{Corollary}
\newcommand{\appnumsection}[1]{\section*{#1}\setcounter{equation}{0}
  \renewcommand{\theequation}{A.\arabic{equation}}
  \renewcommand{\thetheorem}{A.\arabic{theorem}}
  \renewcommand{\thetable}{A.\arabic{table}}
  \renewcommand{\thefigure}{A.\arabic{figure}}
  \renewcommand{\thesection}{A} }
\renewcommand{\theequation}{\arabic{section}.\arabic{equation}}
\renewcommand{\thefootnote}{(\arabic{footnote})}
\newcounter{algo}[section]
\renewcommand{\thealgo}{\thesection.\arabic{algo}}
\newcommand{\llem}[2]{\vspace{\baselineskip} 
\noindent\framebox[\textwidth]{\parbox{0.95\textwidth}{
\begin{lemma} \label{#1} \rm #2 \end{lemma} } } \vspace{\baselineskip} }
\newcommand{\llcor}[2]{\vspace{\baselineskip} 
\noindent\framebox[\textwidth]{\parbox{0.95\textwidth}{
\begin{corollary} \label{#1} \rm #2 \end{corollary} } } \vspace{\baselineskip} }
\newcommand{\algo}[3]{\refstepcounter{algo}
\begin{center}\begin{figure}[htbp]
\framebox[\textwidth]{
\parbox{0.95\textwidth} {\vspace{\topsep}
{\bf Algorithm \thealgo : #2}\label{#1}\\
\vspace*{-\topsep} \mbox{ }\\
{#3} \vspace{\topsep} }}
\end{figure}\end{center}}
\newcommand{\bpr}{{\bf Proof.} \hspace{1.5mm}}
\newcommand{\epr}{\hfill $\Box$ \vspace*{1em}}
\newcommand{\proof}[1]{
\begin{list}{}{
\setlength{\topsep}{0.0pt}
\setlength{\partopsep}{0.0pt}
\setlength{\leftmargin}{0.025\textwidth}
\setlength{\rightmargin}{0.5\leftmargin}
\setlength{\labelwidth}{0.5\leftmargin}
\setlength{\labelsep}{0.25\leftmargin}}
\item \bpr #1 \epr \noindent
\end{list}}
\newcommand{\lthm}[2]{\vspace{\baselineskip} 
\noindent\framebox[\textwidth]{\parbox{0.95\textwidth}{
\begin{theorem} \label{#1} \rm #2 \end{theorem} } } \vspace{\baselineskip} }
\newcommand{\ii}[1]{\{ 1, \ldots, #1 \}}
\newcommand{\iiz}[1]{\{ 0, \ldots, #1 \}}

\renewcommand{\Re}{\hbox{I\hskip -2pt R}}

\newcommand{\bigfrac}[2]{\frac{\displaystyle #1}{\displaystyle #2}}
\newcommand{\bigsum}{\displaystyle \sum}

\newcommand{\sfrac}[2]{{\scriptstyle \frac{#1}{#2}}}
\newcommand{\kap}[1]{\kappa_{\mbox{\tiny #1}}}
\newcommand{\eqdef}{\stackrel{\rm def}{=}}
\newcommand{\tal}[1]{{\normalsize {\sf #1}}}
\newcommand{\half}{\sfrac{1}{2}}

\newcommand{\flow}{f_{\rm low}}

\newcommand{\tG}{\widetilde{G}}
\newcommand{\tg}{\widetilde{g}}
\newcommand{\ip}[1]{\left\langle#1\right\rangle}
\newcommand{\E}[1]{\mathbb{E}\!\left[#1 \right]}
\newcommand{\Econd}[2]{\mathbb{E}_{#1}\!\left[#2 \right]}
\newcommand{\dmax}{d_{\max}}

\newcommand{\diag}{{\rm diag}}

\DeclareMathOperator{\tr}{tr}

\newif\ifcolorcomments
\colorcommentstrue

\newcommand{\private}[1]{}
\newcommand{\comment}[1]{}

\newcommand{\algname}{ADPREC}

\title{A unified convergence theory for adaptive first-order\\ methods
  in the nonconvex case, including  AdaNorm, \\full and diagonal
  AdaGrad and adaptive Muon}

\author{
S. Gratton\thanks{Universit\'{e} de Toulouse, INP, IRIT, Toulouse, France. Email:
     serge.gratton@enseeiht.fr. Work partially supported by 3IA Artificial and
     Natural Intelligence Toulouse Institute (ANITI), French "Investing for the Future
     - PIA3" program under the Grant agreement ANR-19-PI3A-0004"}
~and Ph. L. Toint\thanks{Universit\'{e} de Toulouse, INP, IRIT,
  Toulouse, France, and NAXYS, University of Namur, Namur, Belgium. Email: philippe.toint@unamur.be}
}

\date{\documentdate}
\begin{document}

\renewcommand{\thefootnote}{\fnsymbol{footnote}}
\maketitle
\renewcommand{\thefootnote}{\arabic{footnote}}

\begin{abstract}
A unified framework for first-order optimization algorithms for
nonconvex unconstrained optimization is proposed that uses adaptively
preconditioned gradients and includes popular methods such as full and
diagonal AdaGrad, AdaNorm, as well an adaptive variant of Muon.  This
framework (ADPREC) also allows combining heterogeneous geometries
across different groups of variables while preserving a unified
convergence analysis.  A fully stochastic global rate-of-convergence
analysis is conducted for all methods in the framework, with and
without two types of momentum, using reasonable (although necessarily
stronger than bounded variance) assumptions on the gradient oracle and
without assuming bounded stochastic gradients or small enough
stepsize.
\end{abstract}

{\small
\textbf{Keywords:} Unconstrained nonconvex optimization, first-order
methods, global rate of convergence, Adam, AdaGrad, Muon.
}

\numsection{Introduction}

We consider the problem of finding a first-order critical point for
the optimization problem
\beqn{problem}
\min_{X} f(X)
\eeqn
where $f$ is a smooth, possibly nonconvex function of $X$, whose exact
gradient $G(x) = \nabla_Xf(X)$ is unknown but can be approached by a
random approximation $\tG(X,\xi)$ where $\xi$ is an appropriately defined
random variable.
Stochastic first-order methods for solving problem \req{problem} have
surged as a major research theme in recent years, essentially
motivated by their very successful use in deep learning applications
\cite{BottCurtNoce18}. Because most of these methods do not evaluate
the objective function at all\footnote{They belong to the
Objective-Function-Free Optimization (OFFO) class.}, they are very
robust in the presence of stochastic noise, a crucial feature in these applications.
Starting with stochastic gradient descent \cite{RobbMonr51}, this
domain has seen major advances in the last twenty years, with landmark
proposals such as AdaNorm
\cite{StreMcMa10,DuchHazaSing11,WardWuBott19}, AdaGrad
\cite{DuchHazaSing11,McMaStre10}, Adam \cite{KingBa15}, 
and, more recently, Muon
\cite{Jordetal24}.
The number of their variants (see
\cite{TielHint12,MukkHein17,WuWardBott18,WardWuBott19,AttiKore23,
Vyasetal24,SiZhanShen25} to cite only a few examples)
has grown to the point where a comprehensive review is now a major undertaking (and beyond the
scope of this paper).  Their convergence analysis has followed the
same trend, with significant contributions in particular in
\cite{ZhanXu12,ReddKaleKuma18,WuWardBott18,LiOrab19,DefoBottBachUsun22,
  GuoYinJinYang21,Fawetal22,KaviLevyCevh22,GratJeraToin22b,FawRoutCaraShak23,
  Liuetal23,WangZhanMaChen23,HongLin24,JianMalaMokt24,ZhanZhouZou25,
  LiHong25,Liuetal25,Pethetal25}.

A major theme in this domain has been preconditioning, as a cure to
the known sensitivity of pure gradient methods to (even modest)
problem ill-conditioning. The AdaGrad adaptive diagonal
preconditioning strategy proposed in \cite{StreMcMa10,DuchHazaSing11}
and used in Adam \cite{KingBa15}, has been at the source of many of
the proposals mentioned above, first for unstructured methods
(e.g. \cite{WuWardBott18,GratJeraToin22b,AlacLyu23,AlacMaliCevh21})
and, in the past few years, for methods exploiting the tensor
structure of deep-learning applications
\cite{Vyasetal24,Liuetal25,SiZhanShen25,ZhanLiuScha25}.  The
convergence of many of these methods has been investigated, but is
often specific to the method considered. From our point of view,
notable exceptions are \cite{GuptKoreSing17} whose interesting
interpretation of preconditioning in dual norms covers AdaGrad and
some variants in the convex case, \cite{DefoBottBachUsun22} covering
both Adam and AdaGrad in the nonconvex case, \cite{GratJeraToin22b}
covering AdaGrad and divergent series variants in the nonconvex case,
\cite{Xieetal25,Kova25a} covering AdaGrad, AdaNorm and Shampoo in the
convex case, and \cite{Pethetal25} where the idea of preconditioning
in dual norm is applied to discuss (non-adaptive) variants of Muon in
the nonconvex case.

As it turns out, results for methods for nonconvex problems involving
momentum either establish  non-convergence
\cite{ReddKaleKuma18,Toin23}\footnote{\cite{Toin23} provides a simple example showing
that Adam with fixed stepsize might fail to converge in the deterministic setting,
irrespective of the choice of its parameters, while \cite{ReddKaleKuma18}
gives an example of non-convergence in the stochastic
context. Since our present objective is to analyze complexity of
(convergent) algorithms, we will not consider Adam any further. We
also refer the reader to \cite{ZouShenJieSunLiu19a,DefoBottBachUsun22,Zhanetal22b} for more discussion of this issue.} or
convergence to a mere neighbourhood of a critical point
\cite{DefoBottBachUsun22},  or assume either that gradients are
uniformly bounded \cite{ReddKaleKuma18} or (somewhat unrealistically)
that the user specifies a stepsize parameter that is sufficiently
small compared to the inverse of the problem's Lipschitz constant
\cite{GuoYinJinYang21,HongLin24,XiaoHuLiuToh24},
or to the termination criterion \cite{LiDongLin25,ZhanZhouZou25}.

As far as the authors are aware, the convergence theory for all these
proposals (except \cite{Xieetal25} in the convex case) assumes that the parameters to
optimize are of a single type, either a vector of scalars, or a
matrix, each having its own dedicated preconditioner.  However,
practical applications often mix those types (for instance considering
together activation levels and biases, and weight matrices in neural
networks).
Structured second-order
methods such as K-FAC \cite{MartGros15} approximate curvature
information using block-wise Kronecker factorizations, and these
methods are typically applied only to selected layers, while simpler
and less costly updates are used elsewhere.
This practice is for instance recommended in
\cite{Yuetal17,Ginsetal19,Pethetal25,BernNewh24b}, although without
analysis for adaptive methods. It is therefore
interesting to consider the space of variables as the product space of
blocks of parameters of possibly different types. In particular, this
has the advantage of allowing a unified view of both vectors and
matrix parameters.

Our paper attempts to exploit this observation to propose a
unified algorithmic framework, dubbed \tal{\algname}, and to derive an
associated  general theory.  Its main contributions are the following.
\begin{enumerate}
\item It provides the first (to the authors' knowledge) truly unified
  convergence analysis of adaptive preconditioned gradient methods,
  covering, in the nonconvex case, full and diagonal AdaGrad,
  AdaNorm and adaptive Muon, with and without momentum, and without
  assuming boundedness of the gradients or a sufficiently small stepsize
  parameter.
\item
  It does so by considering a variable/parameter space structured as the
  product of blocks of potentially different types (much as in
  \cite{Xieetal25})
  and by providing new simple proofs based on non-trivial trace inequalities and the theory
  of operator monotone functions. The \tal{\algname} framework allows combining
  heterogeneous geometries across different blocks
  of a Cartesian product space, while preserving a unified and  globally consistent complexity
  analysis. This analysis is conducted under a stochastic assumption
  on the gradient oracle
  that allows the coverage of Muon, which, as we show by an example,
  would be impossible under the standard ``bounded variance'' condition. This unified analysis is of special interest
  for adaptive methods because one may fear that the (possibly
  very) different adaptive stepsizes could generate strong distortions
  between the rates of convergence across blocks. Besides the obvious
  difference in proof techniques, it differs from the
  approach of \cite{Xieetal25} in that it does not require convexity of
  the objective function and covers the case where momentum is used.
\end{enumerate}
  
The paper is organized as follows. Section~\ref{sec:algo} introduces
the structured parameter space and our general algorithmic framework.
Its convergence properties are established in
Section~\ref{sec:theory}, while Section~\ref{sec:appl} details why
this framework covers full and diagonal AdaGrad, AdaNorm and adaptive
Muon.  Some conclusions and perspectives are finally outlined in
Section~\ref{sec:conclusions}.

\noindent
\textbf{Notations:} In what follows, $\|\cdot\|_E$  denotes the
Euclidean norm and the symbols $\succeq$ and  $\preceq$ respectively denote the
``larger or equal'' and ``less or equal'' relations in the L\"{o}wner
semi-order on positive semidefinite matrices. If $\|\cdot\|$ is a norm
on some space $\calS$ endowed with an inner product
$\langle\cdot,\cdot\rangle$, its dual norm $\|\cdot\|_*$ is defined  by
$\|x\|_* = \max_{y\in \calS\setminus\{0\}}\langle y,x \rangle/ \|y\|$. If $x$
  is a vector, $[x]_i$ denotes its $i$-th component.

\numsection{Parameter space structure and algorithm's abstraction}\label{sec:algo}

In order to exploit the possible different types of parameters
occurring in problem \req{problem}, we separate them into $L$ disjoint
blocks of internally uniform type (vectors or matrices), that is
\[
X=(X_1,\dots,X_L),
\qquad
X_\ell= [X]_\ell\in\Re^{n_\ell\times m_\ell}.
\]
Thus each block contains $d_\ell = n_\ell m_\ell$ parameters 
  the same type, but parameter's types may differ between blocks. The
total number of parameters (the dimension of our complete optimization
space) is 
\beqn{dims}
N = \sum_{\ell=1}^L n_\ell m_\ell = \sum_{\ell=1}^Ld_\ell.
\eeqn
We also define $\dmax=\max_{\ell\in \ii{L}} d_\ell$.
Each block $\ell$ is equipped with a norm $\|\cdot\|_\ell$ and its dual norm
$\|\cdot\|_{*,\ell}$. 
The canonical pairing in the Cartesian product space
$\Re^N = \prod_{\ell=1}^L \Re^{d_\ell}$ is then defined as
\[
\ip{U,V} = \sum_{\ell=1}^L \ip{U_\ell,V_\ell}_F,
\tim{ where }
\ip{A,B}_F=\tr(A^T B).
\]
The (primal) product space is endowed with the norm
\[
\|D\|^2=\sum_{\ell=1}^L \|D_\ell\|_\ell^2,
\]
whose dual norm is
\beqn{dual-product-norm}
\|Z\|_*^2=\sum_{\ell=1}^L \|Z_\ell\|_{*,\ell}^2.
\eeqn
(Given a primal norm, the corresponding dual norm
is the natural norm to measure gradients or their approximations. The
use of dual norms for preconditioning has a long history (e.g.\ \cite[Section~9.4]{BoydVand04}),
and has emerged in the machine learning community following 
\cite{Flyn17,GuptKoreSing18,BernNewh24,BernNewh24b,BallPedrLeRo20} or \cite{Pethetal25} for instance.)

In this context, our objective is to find a first-order critical point for problem
\req{problem}, that is an $X_*$ such that
\[
\|G(X_*)\|_* = 0.
\]
In order to define our algorithmic framework to achieve this goal, we
choose, for each block $\ell$, a measurable
normalization operator  $\calN_\ell(\cdot)$  such that,
\[
\calN_\ell(D) \left\{\begin{array}{ll}
 \in \arg\max_{\|V\|_\ell \le 1} \ip{D,V}_F &\tim{if } D\neq 0\\
  = 0 &\tim{if } D=0,
 \end{array}\right.
 \]
which implies that
\beqn{S-normed}
\|\calN_\ell(D)\|_\ell=\left\{\begin{array}{ll} 1 & \tim{if } D \neq 0
\\ 0 &\tim{if }D=0.\end{array}\right.
\tim{ and }
\langle D,\calN_\ell(D)\rangle_F = \|D\|_{*,\ell}.
\eeqn
(A similar definition was used in \cite{GuptKoreSing18} for the
single-block convex case.)
Given that we will study adaptively
preconditioned methods, we finally introduce an operator to construct
it by successive updates.  The operator
$\calU_{k,\ell}(D): \Re^{n_\ell\times  m_\ell}\to S_+^{n_\ell}$ (where $S_+^{n_\ell}$ is the set of
positive semidefinite matrices of size $n_\ell$)
then defines the difference between the preconditioner for block $\ell$ at
iterations $k+1$ and $k$.

Since computing $G_k =\nabla_Xf(X_k)$ is assumed to be either
impossible or too expensive, we consider an iterative stochastic
approach in which, at iterate $X_k$, $G_k$ is approximated by an oracle $\tG_k$
depending on a random variable $\xi_k$.

We now motivate these  concepts with the simple
AdaNorm algorithm \cite{StreMcMa10,DuchHazaSing11,WardWuBott19} on a
single block ($L=1$) in the (self dual) Euclidean norm. This unique
block has dimension $n_1=N$ and $m_1=1$, so $\tG_k$ and $X_k$ are
vectors (now denoted by $\tg_k$ and $x_k$, respectively)

\algo{adanorm}{\tal{AdaNorm}}{
  Given: a starting point $x_0$ and constants $\eta,\varsigma>0$. Set
  $\gamma_{-1}=\varsigma$.\\*[1ex]
  For $k = 0, 1, \ldots$ \\
  \vspace*{-2mm}
  \begin{enumerate}
  \item Draw $\xi_k$ and compute $\tg_k= \tg(x_k,\xi_k) \approx \nabla_Xf(x_k)$,
    \vspace*{-2mm}
  \item $\gamma_k = \gamma_{k-1}+ \|\tg_k\|^2$,  
    \vspace*{-2mm}
  \item $x_{k+1} = x_k - \eta\, \bigfrac{\tg_k}{\sqrt{\gamma_k}}$.
  \end{enumerate}
}  

\noindent
For our purpose, we rewrite the update to $\gamma_k$, now in
matrix form, as
\[
\Gamma_k = \Gamma_{k-1} + \|\tG_k\|^2 I =  \Gamma_{k-1} + \calU_{k,1}(\tG_k),
\]
defining the operator $\calU_{k,1}(D) = \|D\|^2I$ for AdaNorm. The iterate update may also
be reformulated, in its matrix form, as
\[
X_{k+1}
= X_k - \eta\, \bigfrac{\|\tG_k\|}{\sqrt{\Gamma_k}}\cdot\bigfrac{\tG_k}{\|\tG_k\|}
= X_k - \eta\, \|Z_k\|\cdot \calN_1(Z_k)
\]
where
\[
Z_k = \bigfrac{\tG_k}{\sqrt{\gamma_k}}= \Gamma_k^{-1/2}\tG_k
\tim{ and }
\calN_1(Z_k) = \bigfrac{\tG_k}{\|\tG_k\|} = \bigfrac{Z_k}{\|Z_k\|},
\]
and $\calN_1(\cdot)$ is the normalization operator in the Euclidean norm.
Thus the familiar AdaNorm may be reformulated using the $\calU_{k,1}(\cdot)$ and $\calN_1(\cdot)$
operators. We claim (and verify in Section~\ref{sec:appl}) that this is not
only the case for AdaNorm, but the same formalism also applies to
other methods of interest.

Having motivated the introduction of the $\calN_\ell(\cdot)$ and $\calU_{k,\ell}(\cdot)$,
we now state our simple but general algorithmic \tal{\algname} framework
\vpageref{genalgo}.

\algo{genalgo}{\tal{\algname}}{
  Given: a starting point $X_0$, operators $\{\calU_{k,\ell}\}_{k\ge 0,\ell\in\ii{L}}$ and
  $\{\calN_\ell\}_{\ell\in\ii{L}}$  and constants $\eta,\varsigma>0$. Set
  $\Gamma_{-1,\ell}=\varsigma I_\ell$ for $\ell\in\ii{L}$.\\*[1ex]
  For $k = 0, 1, \ldots$ \\
  \vspace*{-2mm}
  \begin{enumerate}
  \item Draw $\xi_k$ and compute $\tG_k = \tG(X_k,\xi_k)=
    (\tG_{k,1},\ldots,\tG_{k,L})\approx G(X_k)$,
    \hspace*{2mm}\mbox{\small [gradient approx.]}
  \item On each block $\ell\in\ii{L}$, compute
    \begin{eqnarray}
    \Gamma_{k,\ell} & =
    &\Gamma_{k-1,\ell}+\calU_{k,\ell}(\tG_{k,\ell}),
    \hspace*{28mm}\mbox{\small [preconditioner update]} \label{Gamma-def}  \\
    Z_{k,\ell}      & = & \Gamma_{k,\ell}^{-1/2}\tG_{k,\ell},
    \hspace*{39.5mm}\mbox{\small [preconditioned gradient]}\label{z-def} \\
    X_{k+1,\ell}    & = & X_{k,\ell} -
    \eta\,\|Z_{k,\ell}\|_{*,\ell}\,\calN_\ell(Z_{k,\ell}).
    \hspace*{16mm}\mbox{\small [scaled normalized step]} \label{xkp1}
    \end{eqnarray}
   \end{enumerate}
}  

We now illustrate these concepts by showing that familiar algorithms
fit in the \tal{\algname} framework. For clarity of exposition,
we use lower case symbols for vectors.

\subsection{AdaNorm \cite{StreMcMa10,DuchHazaSing11,WardWuBott19}}\label{intro:adn}

Taking the AdaNorm  update on a given block $\ell$ corresponds to
choosing  an isotropic scalar geometry in the relevant subspace.
Reformulating the above description for a general block, we see that, if block
$\ell$ has dimension $n_\ell\times 1$, \[
\Gamma_{-1,\ell} = \varsigma I_{n_\ell}
\tim{ and }
\Gamma_{k,\ell}= \Gamma_{k-1,\ell} + \frac{\|\tg_{k,\ell}\|_\ell^2}{n_\ell} I_{n_\ell}
\ms (k\ge 0),
\]
($\tg_{k,\ell}$ is now a vector) so that the contribution of this block to the step is
\[
s_{k,\ell}
=
-\eta \, \Gamma_{k,\ell}^{-1/2}\tg_{k,\ell}
=
-\frac{\eta}{\sqrt{\gamma_{k,\ell}}}\,\tg_{k,\ell}
= - \eta\frac{\tg_{k,\ell}}{\sqrt{\gamma_{k,\ell}}}\frac{\tg_{k,\ell}}{\|\tg_{k,\ell}\|}
\]
when $\Gamma_{k,\ell}=\gamma_{k,\ell}I_{n_\ell}$.
This is exactly the AdaNorm update on block $\ell$.
In \tal{\algname}, this therefore corresponds to choosing
\beqn{adaN}
m_\ell = 1,
\qquad
\|\cdot\|_{*,\ell}=\|\cdot\|_\ell = \|\cdot\|_E,
\qquad
\calN_\ell(\tG_{k,\ell})=\frac{\tG_{k,\ell}}{\|\tG_{k,\ell}\|_\ell},
\tim{ and }
\calU_{k,\ell}(\tG_{k,\ell}) = \frac{\|\tG_{k,\ell}\|_\ell^2}{n_\ell} I_{n_\ell}.
\eeqn

\subsection{Full AdaGrad \cite{DuchHazaSing11}}\label{intro:fadg}

Taking the ``full'' version of AdaGrad \cite{DuchHazaSing11} update on block $\ell$
corresponds to choosing, on a given block $\ell$, a full matrix-valued geometry.
More precisely, if block $\ell$ has dimension $n_\ell$, one considers
\[
\Gamma_{-1,\ell} = \varsigma I_{n_\ell}
\tim{ and }
\Gamma_{k,\ell}= \Gamma_{k-1,\ell} + \tg_{k,\ell}\tg_{k,\ell}^T
\ms (k\ge 0),
\]
so that the contribution of this block to the step is
$
s_{k,\ell} = -\eta \, \Gamma_{k,\ell}^{-1/2}\tg_{k,\ell}.
$
This is exactly the full AdaGrad update on block $\ell$.
In the \tal{\algname} framework, this corresponds to choosing
\beqn{adaF}
m_\ell = 1,
\qquad
\|\cdot\|_{*,\ell}=\|\cdot\|_\ell=\|\cdot\|_E,
\qquad
\calN_\ell(\tG_{k,\ell})=\tG_{k,\ell}/\|\tG_{k,\ell}\|_E,
\tim{ and }
\calU_{k,\ell}(\tG_{k,\ell}) = \tG_{k,\ell}\tG_{k,\ell}^T.
\eeqn

\subsection{Diagonal AdaGrad \cite{DuchHazaSing11,McMaStre10}}\label{intro:cadg}

We now turn to the use of the diagonal/component-wise AdaGrad update on a given block $\ell$.
If block $\ell$ has dimension $n_\ell\times 1$, this geometry is defined by
\[
\Gamma_{-1,\ell} = \varsigma I_{n_\ell}
\tim{ and }
\Gamma_{k,\ell}= \Gamma_{k-1,\ell} + \diag\big([\tg_{k,\ell}]_i^2\big) \ms (k\ge 0)
\tim{ and }
x_{k+1,\ell} = x_{k,\ell} - \eta\Gamma_{k,\ell}^{-1/2}\tg_{k,\ell}.
\]
In our framework, this corresponds to choosing
\beqn{adaD}
m_\ell = 1,
\qquad
\|\cdot\|_{*,\ell}=\|\cdot\|_\ell=\|\cdot\|_E,
\qquad
\calU_{k,\ell}(\tg_{k,\ell})=\diag\big([\tg_{k,\ell}]_i^2\big),
\tim{ and }
\calN_\ell(\tg_{k,\ell})=\frac{\tg_{k,\ell}}{\|\tg_{k,\ell}\|_E},
\eeqn

Observe that we could also view the diagonal AdaGrad algorithm
as a Cartesian product of scalar AdaNorm.
Also note that, as discussed in Section~\ref{sec:momentum},
momentum in the form \req{Mmom}-\req{Zmom} or \req{Gammom2}-\req{Zmom2} may be introduced on this block,
resulting in an \tal{\algname} variant somewhat similar to Adam \cite{KingBa15}.
It is theoretically interesting that this variant enjoys the convergence behaviour
described by Theorem~\ref{thm:conv-mom} and Corollary~\ref{therate2}.

\subsection{Adaptive MUON/AdaGO \cite{Jordetal24,SiZhanShen25,ZhanLiuScha25}}\label{intro:muon}

We now consider an AdaGrad-inspired adaptive version of MUON \cite{Jordetal24},
described as AdaGO for the single block case in \cite{ZhanLiuScha25}. In our framework, this
corresponds to assigning, for each block $\ell$, a geometry adapted to
the structure of neural network parameters. 
We distinguish two types of blocks. For $\ell\in \calM$, $G_{k,\ell}$
is a matrix of size $n_\ell\times m_\ell$, while for
$\ell\in \calA = \ii{L}\setminus \calM$, $G_{k,\ell}$ is a vector in
$\mathbb{R}^{n_\ell}$. 
For $\ell\in\calA$, we use
\[
\|\cdot\|_\ell = \|\cdot\|_E = \|\cdot\|_{*,\ell},
\]
while for $\ell\in\calM$,
\[
\|\cdot\|_\ell = \|\cdot\|_s,
\qquad
\|\cdot\|_{*,\ell}= \|\cdot\|_*,
\]
where $\|\cdot\|_s$ is the spectral norm and $\|\cdot\|_*$ the
  nuclear norm.
The MUON normalization is defined by
\beqn{MUON-N}
\calN_\ell(\tG_{k,\ell}) =
  \left\{\begin{array}{ll}
  U_{k,\ell} V_{k,\ell}^T & \tim{if } \ell\in\calM \tim{and}
  \tG_{k,\ell}=U_{k,\ell}\Sigma_{k,\ell} V_{k,\ell}^T,\\
  \tG_{k,\ell}/\|\tG_{k,\ell}\|_E & \tim{if } \ell\in\calA.
  \end{array}\right.
\eeqn
We then define, for each block $\ell$,
\beqn{MUON-choices}
\Gamma_{k,\ell} = \gamma_{k,\ell}\,I_{d_\ell},
\tim{ where }
\gamma_{-1,\ell} = \varsigma
\tim{ and }
\gamma_{k,\ell}= \left\{\begin{array}{ll}
\gamma_{k-1,\ell} + \|\tG_{k,\ell}\|_*^2 /d_\ell & \tim{if } \ell \in \calM,\\
\gamma_{k-1,\ell} + \|\tG_{k,\ell}\|_E^2/d_\ell  & \tim{if } \ell \in \calA.
\end{array}\right.
\eeqn
This corresponds to choosing, for each block $\ell$,
\[
\calU_{k,\ell}(\tG_{k,\ell}) = \left\{\begin{array}{ll}
\frac{\|\tG_{k,\ell}\|_*^2}{d_\ell} I_{d_\ell} & \tim{if }\ell \in \calM, \\
\frac{\|\tG_{k,\ell}\|_E^2}{d_\ell} I_{d_\ell} & \tim{if }\ell \in \calA.
\end{array}\right.
\]
We also have that $Z_{k,\ell} =\frac{1}{\sqrt{\gamma_{k,\ell}}} \tG_{k,\ell}$.
With these choices, the resulting \tal{\algname} algorithm corresponds to a
block-wise adaptive MUON method, with stepsize $\eta_{\rm muon,\ell} =
\eta/\sqrt{d_\ell}$. 

\subsection{A few comments on \tal{\algname}}

At each iteration,  the random variable $\xi_k$
(whose distribution may depend on $X_k$) is defined on some
probability space $(\Omega,\calF,\mathbb{P})$ 
($\mathbb{E}$ denotes the corresponding expectation). The expectation
conditioned to knowing $(\xi_0,\ldots, \xi_{k-1})$ is denoted by the
symbol $\Econd{k}{\cdot}$.

For each choice of $\calU_{k,\ell}$ and $\calN_\ell$, the
resulting algorithm thus defines a random process
\[
(\tG_k, Z_k, \Gamma_k, X_k),
\]
where $\tG_k= (\tG_{k,1},\ldots, \tG_{k,L})$, $Z_k =
(Z_{k,1},\ldots, Z_{k,L})$,  $\Gamma_k=(\Gamma_{k,1},\ldots,
\Gamma_{k,L})$
and $X_k = (X_{k,1},\ldots,X_{k,L})$.
At first sight, it may seem that the algorithm performs
separate optimization on each block, but this view is deceptive
because interaction between the blocks occurs in the computation of
the approximate gradient at the full-dimensional iterate $X_k$,
  therefore requiring synchronization between blocks.

The name \tal{\algname} alludes to the fact
that $\Gamma_{k,\ell}$ may be interpreted as an adaptive preconditioner for
block $\ell$ and \req{z-def} therefore produces an adaptively preconditioned
(approximate) gradient $Z_{k,\ell}$.

Note that we allow changing the update formula in \req{Gamma-def} for
$\Gamma_{k,\ell}$ at each iteration ($\calU_{k,\ell}$ could
depend on $k$), although, as far as the authors know, this is not
done in practice. We could also consider making the stepsize $\eta$
and/or the initial preconditioner size $\varsigma$ dependent on the block
(that is using $\eta_\ell$ and/or $\varsigma_\ell$), at the cost of
introducing more constants in our already intricate analysis.

\numsection{Convergence analysis}\label{sec:theory}

The convergence analysis for \tal{\algname} hinges on the following standard assumptions.

\begin{assumption}[Boundedness]
\label{ass:bounded-op}
There exists a constant $\flow$ such that $f(X) \ge \flow$ for all
$X$.
\end{assumption}

\begin{assumption}[Smoothness]
\label{ass:smooth-op}
The objective function $f$ is continuously differentiable and has a Lipschitz continuous gradient,
that is there exists a constant $L_G\ge 0$ such that, for all $X,Y\in\Re^N$,
$\| G(X)-G(Y)\|_*\le L_G\|X-Y\|.$
\end{assumption}

\noindent
We now introduce two conditions that define the class of
algorithms within the \tal{\algname} framework, for which we  develop our theory. We present them as
assumptions \emph{on the algorithm(s)}, rather than on the problem.
Although seemingly
abstract, we will demonstrate in Section~\ref{sec:appl} that they do
hold for the methods introduced in Sections~\ref{intro:adn} to \ref{intro:muon}.

\begin{assumption}[Structural identities]
\label{ass:identities}
For all $k\ge 0$ and all $\ell \in \ii{L}$, we have that
\beqn{ineq1}
\|Z_{k,\ell}\|_{*,\ell}\,\ip{\tG_{k,\ell},\calN_\ell(Z_{k,\ell})}_F
= \tr\!\big(\Gamma_{k,\ell}^{-1/2}\calU_{k,\ell}(\tG_{k,\ell})\big),
\eeqn
and
\beqn{ineq2}
\|Z_{k,\ell}\|_{*,\ell}^2=\tr\!\big(\Gamma_{k,\ell}^{-1}\calU_{k,\ell}(\tG_{k,\ell})\big).
\eeqn
\end{assumption}

\noindent
As will become clear in Lemma~\ref{cond-desc} just below, these identities
reformulate the terms associated with first-order descent (for \req{ineq1})
and second-order perturbations (for \req{ineq2}) in terms of
traces. Note that \req{ineq1} and \req{ineq2} are ``iteration-specific''
conditions, in that they only involve $\Gamma_{k,\ell}$
and are independent of the value of $\Gamma_{k-1,\ell}$.

\begin{assumption}[Gradient-preconditioner compatibility]
\label{ass:opttransfer}

There exists a constant $\kappa_\circ > 0$ such
that, for each block $\ell\in\ii{L}$, each $k\ge0$ and  all $G_\ell$,
\begin{equation}
\label{eq:generic}
\|\tG_\ell\|_{*,\ell}^2 \le \kappa_\circ^2 \, \tr\big(\calU_{k,\ell}(\tG_\ell)\big).
\end{equation}
\end{assumption}

\noindent
Because $\calU_{k,\ell}(\cdot)$ is used to build the preconditioners
$\Gamma_{k,\ell}$, this last assumption establishes the link between
them and the measures of (approximate) optimality $\tG_{k,\ell}$. 
(Note that we could choose  $\kappa_\circ$ depending on $\ell$, but we
use a single constant for simplicity.)

We of course need to be more specific on what we assume on the
oracle $\tG_k$, but postpone the precise nature of our requirements to
the statements of the results where they are needed.

Our first step is to analyze the (expected) descent at
iteration $k$.

\llem{cond-desc}{
Suppose that Assumption~\ref{ass:smooth-op} holds. Then
\beqn{descent-k}
\begin{aligned}
\Econd{k}{f(X_{k+1})}
\le & f(X_k)-\eta\sum_{\ell=1}^L\Econd{k}{\|Z_{k,\ell}\|_{*,\ell}\ip{\tG_{k,\ell},\calN_\ell(Z_{k,\ell})}_F}\\
&+\eta\sum_{\ell=1}^L\Econd{k}{\|Z_{k,\ell}\|_{*,\ell}\big|\ip{\tG_{k,\ell}-G_{k,\ell},\calN_\ell(Z_{k,\ell})}_F\big|}
+\frac{L_G\eta^2}{2}\sum_{\ell=1}^L\Econd{k}{\|Z_{k,\ell}\|_{*,\ell}^2}.
\end{aligned}
\eeqn
}
\proof{
By Lipschitz continuity of the gradient (Assumption~\ref{ass:smooth-op}),
\[
f(X_{k+1})
\le f(X_k) +\ip{G_k,X_{k+1}-X_k} + \frac{L_G}{2}\|X_{k+1}-X_k\|^2.
\]
Thus, using \req{z-def} and \req{xkp1}, we obtain
\[
f(X_{k+1})
\le f(X_k)-\eta\sum_{\ell=1}^L\|Z_{k,\ell}\|_{*,\ell}\ip{G_{k,\ell},\calN_\ell(Z_{k,\ell})}_F
   +\frac{L_G\eta^2}{2}\sum_{\ell=1}^L\|Z_{k,\ell}\|_{*,\ell}^2\|\calN_\ell(Z_{k,\ell})\|_\ell^2.
\]
Taking conditional expectations, inserting $\tG_{k,\ell}$ and using
\req{S-normed} yields \req{descent-k}.
} 

\subsection{Trace inequalities}

\noindent
Our argument now takes a little linear algebra detour for proving two useful
inequalities about the trace of matrices of interest.
We first derive an inequality involving the
trace of the inverse square root of the sum of two matrices.

\llem{sqrt-trace}{
Let $A$ be a positive definite matrix and $B$ be a positive
semidefinite matrix. Then
\beqn{sqrt-trace-ineq}
\tr\!\big((A+B)^{-1/2}B\big) \ge \tr((A+B)^{1/2})-\tr(A^{1/2}).
\eeqn
}

\proof{
Set $X = (A+B)^{1/2}$ and $Y = A^{1/2}$.
Clearly, $X \succ 0$, $Y \succ 0$, and $X+Y \succ 0$ and is therefore invertible. Moreover
\beqn{lsq:e1}
B = X^2 - Y^2 = X(X-Y) + (X-Y)Y.
\eeqn
Multiplying this inequality on the right by $(X+Y)^{-1}$ gives
\[
B(X+Y)^{-1}=  X(X-Y)(X+Y)^{-1} + (X-Y)Y(X+Y)^{-1}.
\]
Taking the trace and using cyclicity, we obtain that
\[
\begin{aligned}
\tr\!\big(B(X+Y)^{-1}\big)
&= \tr\!\big((X+Y)^{-1}X(X-Y)\big) + \tr\!\big((X+Y)^{-1}(X-Y)Y\big)\\
&= \tr\!\Big((X+Y)^{-1}\big(X(X-Y)+(X-Y)Y\big)\Big)\\
&= \tr\!\Big((X+Y)^{-1}\big((X-Y)(X+Y)- [X,Y]\big)\Big)\\
\end{aligned}
\]
where $[X,Y] = XY-YX$ is the commutator between $X$ and $Y$.
Since $X$ and $Y$ are Hermitian, $[X,Y]$ is skew-Hermitian and its
trace is zero.
Hence, using cyclicity (twice),
\[
\begin{aligned}
\tr\!\big((X+Y)^{-1}B\big)
= \tr\!\big(B(X+Y)^{-1}\big)
&=\tr\!\big((X+Y)^{-1}(X-Y)(X+Y)\big)-\tr\!\big((X+Y)^{-1}[X,Y]\big)\\
&=\tr\!\big((X+Y)^{-1}(X-Y)(X+Y)\big)\\
&=\tr(X-Y)\\
\end{aligned}
\]
Now, since $Y \succeq 0$, we have that $X + Y \succeq X$.
Since both $X$ and $X+Y$ are positive definite and since inversion
reverses the L\"{o}wner semi-order, we deduce that
\[
(X+Y)^{-1} \preceq X^{-1} = (A+B)^{-1/2}.
\]
and thus $(A+B)^{-1/2}-(X+Y)^{-1}\succeq 0$.
But, since $B$ is symmetric positive semidefinite, we deduce, again using
cyclicity, that
\[
\tr\!\Big( \big((A+B)^{-1/2}-(X+Y)^{-1}\big) B \Big)
= \tr\!\Big( B^{1/2}\big((A+B)^{-1/2}-(X+Y)^{-1}\big) B^{1/2} \Big)
\succeq 0,
\]
and therefore that
\[
\tr\!\Big((A+B)^{1/2}\Big) - \tr\!\Big(A^{1/2}\Big)
= \tr(X-Y)
= \tr\!\Big((X+Y)^{-1}B\Big) \le \tr\!\Big((A+B)^{-1/2}B\Big),
\]
which is \req{sqrt-trace-ineq}.
} 

\noindent
The next step of our argument uses the L\"{o}wner semi-order
$\succeq$ on symmetric positive-semidefinite matrices and the theory
of operator monotone functions (see \cite[Chapter~V]{Bhat96}, for
instance).

\llem{thm:trace-increment-1}{
Let $A$ be a positive definite matrix and $B$ be a positive
semidefinite matrix, and let $\phi:(0,\infty)\to\mathbb R$
be a $C^1$ function. Suppose that $\phi'$ is operator monotone decreasing
on $(0,\infty)$, meaning that for any positive definite matrices
$X,Y$,
$
X\preceq Y \tim{implies that} \phi'(X) \succeq \phi'(Y).
$
Then
\beqn{general-trace-ineq}
\tr\!\big(\phi'(A+B)B\big)
\le \tr\!\big(\phi(A+B)-\phi(A)\big)
\le\tr\!\big(\phi'(A)B\big).
\eeqn
}

\proof{
Define $g(t)\eqdef \tr\!\big(\phi(A+tB)\big)$ for $t\in[0,1]$.
Using the standard derivative formula for spectral functions under the
trace, we see that
\beqn{trderphi}
g'(t)=\tr\!\big(\phi'(A+tB)B\big).
\eeqn
Since $B\succeq0$, we also have that
\beqn{semiorder}
A\preceq A+tB\preceq A+B
\qquad (t\in[0,1]).
\eeqn
and hence, because $\phi'$ is operator monotone decreasing, that
\[
\phi'(A+B)\preceq \phi'(A+tB)\preceq \phi'(A).
\]
Now if $M\preceq N$ and $B\succeq0$, then $\tr((N-M)B)=
\tr(B^{1/2}(N-M)B^{1/2})\ge0$, and thus $\tr(MB)\le\tr(NB)$.
Using this inequality in \req{semiorder} and the identity \req{trderphi}, we
therefore obtain that 
\[
\tr\!\big(\phi'(A+B)B\big)
\le g'(t)
\le \tr\!\big(\phi'(A)B\big).
\]
Integrating over $t\in[0,1]$ then yields \req{general-trace-ineq}.
}

\noindent
We now use the above and a classical result by K. L\"{o}wner
to deduce two helpful inequalities on the trace of matrices generated
by the \tal{\algname} algorithm.

\llem{lem:sqrt-op}{
We have that, for all $k\ge 0$,
\beqn{sqrt-pot}
\sum_{\ell=1}^L \tr(\Gamma_{k,\ell}^{1/2})-\sum_{\ell=1}^L\tr(\Gamma_{-1,\ell}^{1/2})
\le \sum_{j=0}^k\sum_{\ell=1}^L\tr\!\left(\Gamma_{j,\ell}^{-1/2}\calU_{j,\ell}(\tG_{j,\ell})\right)
\eeqn
and
\beqn{log-pot}
\sum_{j=0}^k\sum_{\ell=1}^L\tr\!\left(\Gamma_{j,\ell}^{-1}\calU_{j,\ell}(\tG_{j,\ell})\right)
\le\sum_{\ell=1}^L\tr\!\Big(\log(\Gamma_{k,\ell})\Big)-\sum_{\ell=1}^L\tr\!\Big(\log(\Gamma_{-1,\ell})\Big).
\eeqn
}

\proof{
Let
\beqn{AB-def}
A= \Gamma_{j-1,\ell}
\tim{ and }
B = \calU_{j,\ell}(\tG_{j,\ell}).
\eeqn
Then $\Gamma_{j,\ell} = A+B$ and 
Theorem~\ref{sqrt-trace} gives that, for $j\in\iiz{k}$ and $\ell\in\ii{L}$,
\[
\tr(\Gamma_{j,\ell}^{1/2}) - \tr(\Gamma_{j-1,\ell}^{1/2})
\le \tr\!\Big(\Gamma_{j,\ell}^{-1/2}\calU_{j,\ell}(\tG_{j,\ell})\Big).
\]
Summing this inequality for $j\in\iiz{k}$ and $\ell\in\ii{L}$ yields \req{sqrt-pot}.
In order to prove \req{log-pot}, let $\phi(t)=\log t$. Then
$
\phi'(t)=\frac{1}{t}.
$
which is operator monotone decreasing (see
\cite{Loew34} as cited in \cite[p. 149]{Bhat96}).
Applying the first inequality of \req{general-trace-ineq} in
Theorem~\ref{thm:trace-increment-1} then yields that 
\[
\tr\!\Big((A+B)^{-1}B\Big) \le \tr\!\Big(\log(A+B)-\log A\Big),
\]
which, with \req{AB-def}, ensures that, for $j\in\iiz{k}$ and $\ell\in\ii{L}$,
\[
\tr\!\Big(\Gamma_{j,\ell}^{-1}\calU_{j,\ell}(\tG_{j,\ell})\Big)
\le \tr\!\Big(\log(\Gamma_{j,\ell})\Big) - \tr\!\Big(\log(\Gamma_{j-1,\ell})\Big).
\]
Summing this inequality for $j\in\iiz{k}$ and $\ell\in\ii{L}$ now yields \req{log-pot}.
} 

\subsection{Telescoping inequality and global rate of convergence}

We next introduce a variant of the classical telescoping argument,
leading to the central inequality in our analysis.

\lthm{th:master-op}{
Suppose that Assumption~\ref{ass:bounded-op} and \ref{ass:identities}  hold.
Suppose also that, for some $\omega\ge0$ and $\nu_k \ge0$
\beqn{var-cond}
\sum_{j=0}^k \E{\|\tG_j-G_j\|_*^2}
\le \nu_k^2 + \omega^2\sum_{j=0}^k\sum_{\ell=1}^L\E{\|Z_{j,\ell}\|_{*,\ell}^2}.
\eeqn
Define
\beqn{Delta-def}
\Delta_k
\eqdef \E{\sum_{\ell=1}^L \tr(\log\Gamma_{k,\ell})-\sum_{\ell=1}^L \tr(\log\Gamma_{-1,\ell})}.
\eeqn
Then, for every $k\ge0$,
\beqn{master}
\E{\sum_{\ell=1}^L \tr(\Gamma_{k,\ell}^{1/2})}
\le\kap{gap}+\nu_k\,\sqrt{\Delta_k}
+\left(\omega+\frac{L_G\eta}{2}\right)\,\Delta_k,
\eeqn
where
\beqn{kapgap-def}
\kap{gap} \eqdef \frac{f(X_0)-f_{\rm low}}{\eta} + \sqrt{\varsigma}\,N.
\eeqn
}

\proof{
Taking the full expectation in the conditional descent inequality
\req{descent-k} and summing for $j=0$ to $k$ gives
\beqn{descent-more}
\begin{aligned}
\eta\sum_{j=0}^k\sum_{\ell=1}^L&\E{\|Z_{j,\ell}\|_{*,\ell}\,\ip{\tG_{j,\ell},\calN_\ell(Z_{j,\ell})}_F}
\le f(X_0)-f_{\rm low}\\
&+\eta\sum_{j=0}^k\sum_{\ell=1}^L\E{\|Z_{j,\ell}\|_{*,\ell}\,\big|\ip{\tG_{j,\ell}-G_{j,\ell},\calN_\ell(Z_{j,\ell})}_F\big|}
+\frac{L_G\eta^2}{2}\sum_{j=0}^k\sum_{\ell=1}^L\E{\|Z_{j,\ell}\|_{*,\ell}^2}.
\end{aligned}
\eeqn
Using successively \req{ineq1} and \req{sqrt-pot}, we obtain that
\beqn{lin-term}
\begin{aligned}
\eta \sum_{j=0}^k \sum_{\ell=1}^L\E{\|Z_{j,\ell}\|_{*,\ell}\,\ip{\tG_{j,\ell},\calN_\ell(Z_{j,\ell})}_F}
&=\eta\,\sum_{j=0}^k\sum_{\ell=1}^L\E{\tr\!\bigl(\Gamma_{j,\ell}^{-1/2}\calU_{j,\ell}(\tG_{j,\ell})\bigr)}\\
&\ge \eta\,\E{\sum_{\ell=1}^L
  \tr(\Gamma_{k,\ell}^{1/2})-\sum_{\ell=1}^L
  \tr(\Gamma_{-1,\ell}^{1/2})}.
\end{aligned}
\eeqn
Now observe that the definition of the dual norm and \req{S-normed} give that
\[
\big|
\ip{\tG_{j,\ell}-G_{j,\ell},\calN_\ell(Z_{j,\ell})}_F
\big|
\le
\|\tG_{j,\ell}-G_{j,\ell}\|_{*,\ell}\,
\|\calN_\ell(Z_{j,\ell})\|_\ell
\le \|\tG_{j,\ell}-G_{j,\ell}\|_{*,\ell}.
\]
Therefore
\[
\E{\|Z_{j,\ell}\|_{*,\ell}\,\big|\ip{\tG_{j,\ell}-G_{j,\ell},\calN_\ell(Z_{j,\ell})}_F\big|}
\le \sqrt{\E{\|\tG_{j,\ell}-G_{j,\ell}\|_{*,\ell}^2}}\,\sqrt{\E{\|Z_{j,\ell}\|_{*,\ell}^2}}.
\]
Observe now that the Cauchy-Schwartz inequality also implies that, for
any vectors $a$ and $b$ with nonnegative components,
\beqn{CSsqrt}
\sum_j\sqrt{a_j}\sqrt{b_j}
\le \|\sqrt{a}\|_E\,\|\sqrt{b}\|_E
= \sqrt{\sum_j a_j} \sqrt{\sum_j b_j}.
\eeqn
Hence, summing over $\ell$ and using \req{dual-product-norm}, we obtain that
\[
\sum_{\ell=1}^L\E{\|Z_{j,\ell}\|_{*,\ell}\,\big|\ip{\tG_{j,\ell}-G_{j,\ell},\calN_\ell(Z_{j,\ell})}_F\big|}
\le \sqrt{\E{\|\tG_j-G_j\|_*^2}}\,\sqrt{\sum_{\ell=1}^L\E{\|Z_{j,\ell}\|_{*,\ell}^2}}.
\]
Summing now over $j$, using $\sqrt{a+b}\le\sqrt{a}+\sqrt{b}$ and \req{CSsqrt} again, we
deduce
that
\[
\begin{aligned}
\eta\sum_{j=0}^k\sum_{\ell=1}^L&\E{\|Z_{j,\ell}\|_{*,\ell}\,\big|\ip{\tG_{j,\ell}-G_{j,\ell},\calN_\ell(Z_{j,\ell})}_F\big|}\\
&\le\eta\,\sqrt{\sum_{j=0}^k\E{\|\tG_j-G_j\|_*^2}}\,
          \sqrt{\sum_{j=0}^k\sum_{\ell=1}^L\E{\|Z_{j,\ell}\|_{*,\ell}^2}}\\
& \le \eta \nu_k
          \sqrt{\sum_{j=0}^k\sum_{\ell=1}^L\E{\|Z_{j,\ell}\|_{*,\ell}^2}}
          + \eta\omega
          \sum_{j=0}^k\sum_{\ell=1}^L\E{\|Z_{j,\ell}\|_{*,\ell}^2}
\end{aligned}          
\]
where we used \req{var-cond} to obtain the last inequality.
But, by \req{ineq2}, \req{log-pot} and \req{Delta-def},
\beqn{Delta-low}
\sum_{j=0}^k\sum_{\ell=1}^L\E{\|Z_{j,\ell}\|_{*,\ell}^2}
=\sum_{j=0}^k\sum_{\ell=1}^L\E{\tr\!\bigl(\Gamma_{j,\ell}^{-1}\calU_{j,\ell}(\tG_{j,\ell})\bigr)}
\le \Delta_k,
\eeqn
so that
\beqn{noise-term}
\eta\sum_{j=0}^k\sum_{\ell=1}^L\E{\|Z_{j,\ell}\|_{*,\ell}\,\big|\ip{\tG_{j,\ell}-G_{j,\ell},\calN_\ell(Z_{j,\ell})}_F\big|}
\le \eta\,\nu_k\,\sqrt{\Delta_k} + \eta \omega \Delta_k.
\eeqn
Similarly, using \req{S-normed} and \req{Delta-low}, the quadratic term satisfies
\beqn{q-term}
\frac{L_G\eta^2}{2}\sum_{j=0}^k\sum_{\ell=1}^L\E{\|Z_{j,\ell}\|_{*,\ell}^2\|\calN_\ell(Z_{j,\ell})\|_\ell^2}
=\frac{L_G\eta^2}{2}\sum_{j=0}^k\sum_{\ell=1}^L\E{\|Z_{j,\ell}\|_{*,\ell}^2}
\le \frac{L_G\eta^2}{2}\Delta_k.
\eeqn
Substituting \req{lin-term}, \req{noise-term} and \req{q-term} into
\req{descent-more} then yields that
\beqn{telesc-1}
\eta\,\E{\sum_{\ell=1}^L \tr(\Gamma_{k,\ell}^{1/2})-\sum_{\ell=1}^L \tr(\Gamma_{-1,\ell}^{1/2})}
\le f(X_0)-f_{\rm low} 
+ \eta\,\nu_k\,\sqrt{\Delta_k}
+ \left(\eta \omega +\frac{L_G\eta^2}{2}\right)\Delta_k,
\eeqn
which is exactly \req{master}-\req{kapgap-def} after taking into
account that $\Gamma_{-1,\ell}=\varsigma I_\ell$ for all $\ell\in\ii{L}$.
}

\noindent
Observe that the requirement \req{var-cond} defines an upper bound on
the \textit{cumulative} total variance of the gradient oracle over all past
iterates, an approach more general than assuming conditional variance
at every iteration. In particular, it allows large variance at early
iterations provided later iterations compensate.  Trade-offs between
different realizations are also theoretically possible.
Also note that, since we will prove that $\Gamma_k$ grows very slowly
(see \req{Thetak-def} below), and $Z_k$ is the preconditioned
approximate gradient, condition \req{var-cond} is akin to a cumulative
affine variance condition of the form
\[
\sum_{j=0}^k\sum_{\ell=1}^L \Econd{j}{\|\tG_{j,\ell}-G_{j,\ell}\|_{*,\ell}^2}
\le \nu_k^2 +
\omega^2\sum_{j=0}^k\sum_{\ell=1}^L\Econd{j}{\|G_{j,\ell}\|_{*,\ell}^2},
\]
whose iteration-wise variant has already been used in analysis of
first-order methods (see \cite{Fawetal22,WangZhanMaChen23} or, for the
stronger `` affine-$^*$ '' version, \cite{AttiKore23}). In particular, the ``strong growth''
assumption motivated by over-parametrized problems and used in
\cite{WangZhanMaChen23} to derive an improved 
convergence rate for AdaGrad is essentially subsumed
(at the cumulative level) if ``preconditioned cumulative strong growth''
(that is $\nu_k=0$) is assumed in \req{var-cond}.

We now start exploiting this result by deriving a more specific bound
on the value of $\Delta_k$ in \req{Delta-def}. This requires the
following simple technical lemma.

\llem{lem:spectral-log-square}{
Let $\Gamma\in\Re^{d\times d}$ be symmetric positive definite.
Then
\[
\tr(\log\Gamma) \le 2d\log\!\bigl(\tr(\Gamma^{1/2})\bigr)-d\log(d).
\]
}

\proof{
Let $\lambda_1,\dots,\lambda_d>0$ be the eigenvalues of $\Gamma$.
Then the concavity of the logarithm ensures that
\[
\tr(\log\Gamma)
=\sum_{i=1}^d \log(\lambda_i)
\le d \log\!\left(\frac{1}{d}\sum_{i=1}^d \lambda_i\right)
\le d\log\!\left(\frac{\left(\sum_{i=1}^d \sqrt{\lambda_i}\right)^2}{d}\right)
=  2d\log\!\bigl(\tr(\Gamma^{1/2})\bigr)-d\log(d).
\]
}

\noindent
We now derive the desired bound on $\Delta_k$ in \req{Delta-def}.

\llem{lem:delta-closure}{
We have that, for all $k\ge0$,
\beqn{Delta-upper}
\Delta_k\le \kappa_d + 2N\log\!\left(\E{\sum_{\ell=1}^L \tr(\Gamma_{k,\ell}^{1/2})}\right),
\eeqn
\vspace*{-2mm}
where
\beqn{kaps-def}
\kappa_d = -\sum_{\ell=1}^L d_\ell\log(d_\ell) - \log(\varsigma) N.
\eeqn
}

\proof{
Applying Lemma~\ref{lem:spectral-log-square} to each block gives
\beqn{lem:e1}
\sum_{\ell=1}^L \tr(\log\Gamma_{k,\ell})
\le 2\sum_{\ell=1}^L d_\ell\log\!\bigl(\tr(\Gamma_{k,\ell}^{1/2})\bigr)
-\sum_{\ell=1}^L d_\ell\log d_\ell.
\eeqn
Since
$\tr(\Gamma_{k,\ell}^{1/2}) \le \bigsum_{\ell=1}^L \tr(\Gamma_{k,\ell}^{1/2}),$
we have that
$\log\!\bigl(\tr(\Gamma_{k,\ell}^{1/2})\bigr)
\le \log\!\left(\bigsum_{\ell=1}^L \tr(\Gamma_{k,\ell}^{1/2})\right),$
and hence, from \req{lem:e1},
\[
\sum_{\ell=1}^L \tr(\log\Gamma_{k,\ell})
\le 2\sum_{\ell=1}^L d_\ell\log\!\left(\sum_{\ell=1}^L \tr(\Gamma_{k,\ell}^{1/2})\right)
-\sum_{\ell=1}^L d_\ell\log d_\ell.
\]
We therefore obtain from \req{Delta-def} that 
\beqn{lem:e2}
\Delta_k
\le -\sum_{\ell=1}^L d_\ell\log d_\ell  -\sum_{\ell=1}^L \tr(\log\Gamma_{-1,\ell})
    +2\sum_{\ell=1}^L d_\ell\,\E{\log\!\left(\sum_{\ell=1}^L \tr(\Gamma_{k,\ell}^{1/2})\right)}.
\eeqn
But Jensen's inequality for the concave $\log(\cdot)$ function gives that
\[
\E{\log\!\left(\sum_{\ell=1}^L \tr(\Gamma_{k,\ell}^{1/2})\right)}
\le \log\!\left(\E{\sum_{\ell=1}^L \tr(\Gamma_{k,\ell}^{1/2})}\right).
\]
Inserting this inequality in \req{lem:e2} and using the definition
$\Gamma_{-1,\ell}=\varsigma I_\ell$ gives \req{Delta-upper}-\req{kaps-def}.
}

\noindent
We still need a simple technical result before concluding.

\llem{techn}{Suppose that $1 \le t \le c\log(t)$ for $c>0$. Then
$t< 2c\log(2c)$.
}

  \proof{The concavity of the logarithm at $2$ gives that, for all $x>0$
  \[
  \log(x)\le \log(2)+\frac{x-2}{2} =\frac{x}{2}+\log(2)-1.
  \]
  Applying this inequality for $x = t/c$ then gives that
  \private{
  \[
  \log\left(\frac{t}{c}\right)\le \frac{t}{2c} + \log(2) -1
  \]
  \[
  \log(t) 
  \le \frac{t}{2c} + \log\left(c\right) + \log(2) -1
  \]
  }
  \[
  \log(t)\le\frac{t}{2c} +\log\left(2c\right)-1
  < \frac{t}{2c} +\log\left(2c\right),
  \]
  which we may substitute  in the inequality $2t\le 2c\log(t)$, yielding that
  \[
  2t
  < 2c\left(\frac{t}{2c} + \log\left(2c\right)\right)
   =  t + 2c\log\left(2c\right),
  \]
  and the result follows by substracting $t$ from both sides.
}

\noindent
We now state a first important result on the expected size of the
preconditioners $\Gamma_{k,\ell}$.

\lthm{thm:convergence-theta}{
Suppose that the \tal{\algname} algorithm is applied to problem \req{problem}.
Suppose that \req{var-cond} and Assumptions~\ref{ass:bounded-op},
\ref{ass:smooth-op} and \ref{ass:identities} hold.
Then, for all $k \ge 0$,
\beqn{Thetak-def}
\E{\sum_{\ell=1}^L \tr(\Gamma_{k,\ell}^{1/2}) }  \le \Theta_k
\eqdef
\max\left[\,e^{\max\left[1,\frac{\kappa_d}{2N}\right]},
  \,3\kap{gap}, T_k, Y\,\right],
\eeqn
with $\kap{gap}$ defined in \req{kapgap-def},
\[
T_k =12\sqrt{N}\,
\nu_k\,\sqrt{\max\left[1,\log\left(12\sqrt{N}\,\nu_k\right)\right]},
\ms
Y =
24N\left(\omega+\frac{L_G\eta}{2}\right)
\log\left(24N\left(\omega+\frac{L_G\eta}{2}\right)\right)
\]
$\nu_k$ being defined in \req{var-cond}
and $\kappa_d$ in \req{kaps-def}.
}

\proof{
Define
\beqn{tk-def}
t_k = \E{\sum_{\ell=1}^L \tr(\Gamma_{k,\ell}^{1/2}) }.
\eeqn
Substituting \req{Delta-upper} into \req{master} and using this
definition then gives that
\beqn{th:e1}
t_k
\le \kappa_{\rm gap} +\nu_k\,
     \sqrt{\kappa_d+2N\log\left(t_k\right)}
   + \left(\omega+\frac{L_G\eta}{2}\right)\bigl(\kappa_d+2N\log\left(t_k\right)\bigr).
\eeqn
Suppose first that
\[
\log\left(t_k\right)
< \max\left[\frac{1}{2N},\frac{\kappa_d}{2N}\right].
\]
Then
\beqn{simplecase}
t_k
\le e^{ \max\left[\frac{1}{2N},\frac{\kappa_d}{2N}\right]}.
\eeqn
Alternatively, suppose now that
\beqn{notsimple}
2N\log\left(t_k\right)
\ge \max[1,\kappa_d].
\eeqn
Then
\[
\kappa_d+2N\log\left(t_k\right)
\le 4N\log\left(t_k\right)
\]
and \req{th:e1} can then be rewritten as
\beqn{ineqtk}
t_k \le a + b\,\nu_k\sqrt{\log(t_k)} + c \log(t_k)
\eeqn
with
\beqn{abc-def}
a = \kap{gap},
\ms
b = 2\sqrt{N}
\tim{ and }
c =4N \left(\omega+\frac{L_G\eta}{2}\right).
\eeqn
This formulation is analyzed by distinguishing three cases.\\
$\bullet$ The first case is when $\max[a, b\nu_k\sqrt{\log(t_k)} ,c \log(t_k)]=a$.
Then, clearly,
\beqn{bound-case1}
t_k \le 3a.
\eeqn
$\bullet$ The second case is when
\beqn{case2}
\max[a, b\nu_k\sqrt{\log(t_k)} ,c\log(t_k)]=b\nu_k\sqrt{\log(t_k)}.
\eeqn
If $\log(t_k) \le 1$, then $t_k \le e$.
Alternatively, if $\log(t_k) > 1$ (implying that $t_k>1$), we deduce that
$
t_k^2  \le 9b^2\nu_k^2\log(t_k),
$
that is
\beqn{htk}
h_k(t_k) \eqdef \frac{t_k^2}{\tau_k^2} - \log(t_k) \le 0
\tim{ with } \tau_k = 3b \nu_k.
\eeqn
Let $T_k= 2\tau_k \sqrt{\max[1,\log(2\tau_k)]}\ge 2\tau_k$. Then
\beqn{hTk}
h_k(T_k)
= 4\max[1,\log(2\tau_k)]-\log(2\tau_k) -
\frac{1}{2}\log(\max[1,\log(2\tau_k)])
>0.
\eeqn
Moreover,
$h_k'(t) = \bigfrac{2t}{\tau_k^2} - \bigfrac{1}{t}$,
and therefore, for all $t \ge 2\tau_k$,
$
h_k'(t) \ge \bigfrac{4\tau_k}{\tau_k^2}-\bigfrac{1}{2\tau_k}
= \bigfrac{4}{\tau_k} -\bigfrac{1}{2\tau_k} >0.
$
As a consequence, $h_k(t)$ is increasing for $t\ge 2\tau_k$ and
thus $h_k(t) > 0$ for $t\ge T_k$ because of \req{hTk}. We may then
deduce from \req{htk} that $t_k < T_k$.
Thus,
\beqn{bound-case2}
t_k \le \max\left[e,T_k\right].
\eeqn
$\bullet$ Finally, the third case is when $\max[a, b\nu_k\sqrt{\log(t_k)} ,c \log(t_k)]=c \log(t_k)$.
Then $t_k \le 3c\log(t_k)$ and thus, using Lemma~\ref{techn},
\beqn{bound-case3}
t_k\le \max[ 1, 6c\log(6c)].
\eeqn
Combining \req{tk-def}, \req{bound-case1}, \req{bound-case2} and \req{bound-case3}
and including \req{simplecase} then gives
\req{Thetak-def}.
} 

\noindent
This result finally gives us all the ingredients to derive convergence
and complexity results on the Cartesian product space.  The bound
\req{Thetak-def}, together with Assumption~\ref{ass:opttransfer},
indeed implies convergence of the criticality measure on each block
with a {\em uniform rate-of-convergence bound}.

\lthm{thm:convergence}{
Suppose that the \tal{\algname} algorithm is applied to problem \req{problem}.
Suppose that \req{var-cond} and Assumptions~\ref{ass:bounded-op} to \ref{ass:opttransfer}
hold. Then
\vspace*{-2mm}
\beqn{therate}
\sum_{j=0}^k \E{\|\tG_j\|_*}
\le \kappa_\circ\,\sqrt{k+1}\,\Theta_k
\eeqn
where $\Theta_k$ is defined in Theorem~\ref{thm:convergence-theta}.
Moreover, if $\tG_k$ is an unbiased estimator of $G_k$,
i.e. $\Econd{k}{\tG_k}= G_k$ for all $k\ge 0$, then
\vspace*{-2mm}
\beqn{thetruerate}
\min_{j\in\iiz{k}} \E{\|G_j\|_*}
\le \frac{1}{k+1}\sum_{j=0}^k \E{\|G_j\|_*}
\le \frac{\kappa_\circ\,\Theta_k}{\sqrt{k+1}}.
\eeqn
}

\proof{
The Cauchy-Schwarz inequality and Assumption~4 imply that
\[
\begin{aligned}
\sum_{j=0}^k \|\tG_j\|_*
&\le \sqrt{k+1} \left(\sum_{j=0}^k \|\tG_j\|_*^2\right)^{1/2}\\
&= \sqrt{k+1} \left(\sum_{j=0}^k  \sum_{\ell=1}^L \|\tG_{j,\ell}\|_{*,\ell}^2\right)^{1/2}\\
&\le \kappa_\circ\,\sqrt{k+1} \,
\left( \sum_{j=0}^k\sum_{\ell=1}^L\tr\bigl(\calU_{j,\ell}(\tG_{j,\ell})\bigr)\right)^{1/2}.
\end{aligned}
\]
Now, \req{Gamma-def} ensures that
\[
\sum_{j=0}^k\calU_{j,\ell}(\tG_{j,\ell})
= \Gamma_{k,\ell}-\Gamma_{-1,\ell},
\]
so that
\[
\sum_{j=0}^k \|\tG_j\|_*
\le \kappa_\circ \,\sqrt{k+1}\,\left( \sum_{\ell=1}^L \tr(\Gamma_{k,\ell})\right)^{1/2}
\le \kappa_\circ \,\sqrt{k+1}\, \sum_{\ell=1}^L \tr(\Gamma_{k,\ell}^{1/2}).
\]
Taking the total expectation on both sides of this inequality and using
Theorem~\ref{thm:convergence-theta} then yields that
\[
\sum_{j=0}^k \E{\|\tG_j\|_*}
\le \kappa_\circ\,\sqrt{k+1}\,\E{\sum_{\ell=1}^L \tr(\Gamma_{k,\ell}^{1/2})}
\le \kappa_\circ\,\sqrt{k+1}\,\Theta_k,
\]
proving \req{therate}.
Dividing by $k+1$ and using $\Econd{k}{\tG_k}= G_k$ together with Jensen's inequality gives \req{thetruerate}.
} 

\noindent
We now investigate what can be said if one makes a more specific
assumption on conditional variance at each iteration.

\llcor{therate2}{
Suppose that the \tal{\algname} algorithm is applied to problem \req{problem}.
Suppose that Assumptions~\ref{ass:bounded-op} to \ref{ass:opttransfer}
hold. Suppose also that
\beqn{betaisktogamma}
\Econd{k}{\tG_k}= G_k
\tim{ and }
\Econd{k}{\|\tG_{k,\ell}-G_{k,\ell}\|_{*,\ell}^2}
\le \frac{\sigma_\ell^2}{(k+1)^\alpha}+\omega^2\Econd{k}{\|Z_{k,\ell}\|_{*,\ell}^2}
\eeqn
for some $\alpha,\omega >0$ and all $k \ge 0$ and $\ell\in\ii{L}$.  Then
\beqn{detailed-rates}
\frac{1}{k+1}\sum_{j=0}^k \E{\|G_j\|_*}=
\left\{\begin{array}{ll}
\calO\left(\bigfrac{\sqrt{\log(k+1)}}{(k+1)^{\sfrac{\alpha}{2}}}\right)
&\tim{if } \alpha<1,\\
\calO\left(\bigfrac{\sqrt{\log(k+1)\log(\log(k+1))}}{\sqrt{k+1}}\right)
&\tim{if } \alpha=1,\\
\calO\left(\bigfrac{1}{\sqrt{k+1}}\right)
&\tim{if } \alpha>1.
\end{array}\right.
\eeqn
}

\proof{
Observe first that the term in the expression of $\Theta_k$ in
\req{Thetak-def} with potential fastest growth when
$k$ tends to infinity is $T_k$, yielding (in order)
\beqn{Theta-nomg}
\Theta_k = \calO\left(\nu_k\sqrt{\log(\nu_k)}\right).
\eeqn
But \req{var-cond} and \req{betaisktogamma} imply that
\[
\nu_k^2 = \sigma_{\rm tot}^2 \sum_{j=0}^k\frac{1}{(j+1)^\alpha},
\]
where $\sigma_{\rm tot}^2 = \sum_{\ell=1}^L \sigma_\ell^2$.
Suppose first that $\alpha<1$. We then have that
$\sum_{j=0}^k\frac{1}{(j+1)^\alpha}\le (k+1)^{1-\alpha}/(1-\alpha)$
so that
$\nu_k=\frac{\sigma_{\rm
    tot}}{\sqrt{1-\alpha}}\,(k+1)^{\half(1-\alpha)}$ in this case.
If $\alpha=1$, then
$
\sum_{j=0}^k\frac{1}{(j+1)^\alpha}= \sum_{j=0}^k\frac{1}{j+1} \le
1+\log(k+1)
$
and thus $\nu_k\le \sigma_{\rm tot}\sqrt{1+\log(k+1)}$.
Finally, if $\alpha > 1$, we have that
$
\sum_{j=0}^k\frac{1}{(j+1)^\alpha} \le \zeta(\alpha)<+\infty,
$
where $\zeta(\cdot)$ is the Riemann zeta function.
Combining the three cases, we obtain that
\beqn{nuk-vals}
\nu_k \le \left\{\begin{array}{ll}
\bigfrac{\sigma_{\rm tot}}{\sqrt{1-\alpha}}(k+1)^{\half(1-\alpha)}\hspace*{20mm}
&\tim{if }\alpha<1,\\
\sigma_{\rm tot}\sqrt{1+\log(k+1)}
&\tim{if }\alpha=1,\\*[1.5ex]
\sigma_{\rm tot}\zeta(\alpha)
&\tim{if }\alpha>1,\\
\end{array}\right.
\eeqn
The rates \req{detailed-rates} then follow from these bounds and
\req{thetruerate}.
}

\noindent
Note that the continuity of the bound expressed in Theorem~\ref{thm:convergence}  as
a function of $\nu_k$ is lost in the statement of
Corollary~\ref{therate2}, because, as is clear from its proof, the
constants hidden in the $\calO(\cdot)$ notation depend on $\alpha$. In
particular, the formulation using $\calO(\cdot)$ does not allow taking
the limit for $\alpha$ tending to one.

Observe that we could replace the second part of \req{betaisktogamma} by
\beqn{alt-var-cond}
\Econd{k}{\|\tG_{k,\ell}-G_{k,\ell}\|_{*,\ell}^2}
\le \frac{\sigma_\ell^2}{(k+1)^\alpha}+\omega^2\|Z_{k-1,\ell}\|_{*,\ell}^2
\eeqn
with the convention that $Z_{-1,\ell}=0$, since
\[
\sum_{j=0}^{k-1}\sum_{\ell=1}^L\Econd{j}{\|Z_{j,\ell}\|_{*,\ell}^2}
\le \sum_{j=0}^k\sum_{\ell=1}^L\Econd{j}{\|Z_{j,\ell}\|_{*,\ell}^2}.
\]
The advantage of \req{alt-var-cond}  is that the right-hand-side of this new condition
is now  $\calF_k$-measurable at iteration $k$ while this is not the case for the second part of
\req{betaisktogamma} because the $Z_{k,\ell}$ are not $\calF_k$-measurable.

Although the rates of convergence obtained in Corollary~\ref{therate2} under the
bias and variance conditions \req{betaisktogamma} are reasonable, they
do not quite match, for high-variance regimes, those obtained for
(momentum-less) AdaGrad and AdaNorm \cite{WangZhanMaChen23}, which do
not depend on the finiteness of the ``inaccuracy
budget'' $\nu_k^2$ and thus allow for bounded conditional variance at
every iteration (condition \req{var-cond} may require increasing batch-size) .  However
they recover (in order) the best\footnote{It is not known
  whether a better rate is achievable for AdaGrad and Adanorm.}
$\calO(\sqrt{\log(k+1)\log(\log(k+1))/(k+1)})$ rate obtained in
\cite{WangZhanMaChen23} in the
preconditioned cumulative strong growth 
context\footnote{That is not only in the case where $\sigma_\ell = 0$ for each $\ell$,
but, more generally, in the case where the ``inaccuracy
budget'' $\nu_k^2$  is finite.}. Importantly, they do so for a large
class of algorithms.
While it would be desirable to derive general convergence
  results under the standard ``bounded variance'' assumption (that is
  $\alpha=\omega=0$ in \req{betaisktogamma}), this turns out to be impossible for a unified
  theory covering Muon, as shown by the following simple example.
  Consider using Muon (see Section\req{intro:muon}) to minimize
  \[
  f(X) = \frac{1}{4}\big([X]_{1,1}-[X]_{2,2}\big)^2
  \tim{ starting from } X_0 = \diag( 1, -1) \in \Re^{2\times2}.
  \]
  We have that $G(X) = \diag(r(X),-r(X))$ with
  $r(X)=\half([X]_{1,1}-[X]_{2,2})$, so that $r(X_0)= 2$ and $\|G(X_0)\|_*=1$ where
  $\|\cdot\|_*$ is the nuclear norm. At iteration $k$, let
  $\tG_k = \diag(3\,r(X_k), r(X_k))$ or $\tG(X) = \diag(-r(X_k), 3\,r(X_k))$
  with equal probabilities. Then $\Econd{k}{\tG_k}=G_k$,
  $\Econd{k}{\|\tG_k-G_k\|_*^2} = 16\, r(X_k)^2$, and one verifies that
  $\calN_1(\tG_k)$ given by \req{MUON-N} is then either the identity or its opposite with
  equal probabilities.  Thus every Muon step is a multiple of the
  identity matrix, which leaves $r(X)$, and therefore $G(X)$,
  unchanged. As a consequence, $\|G(X_k)\|_*=1$ for all $k\ge 0$,
  despite  $\Econd{k}{\|\tG_k-G_k\|_*^2}=64$ remaining bounded for all $k$.
We also note that, should the restrictive
assumption of a bounded gradient oracle be made for AdaGrad, only the
first part of \req{betaisktogamma} (unbiased oracle) is necessary to
deduce that the average of $\E{\|G_k\|_E^2}$ decreases 
in $\calO(1/(k+1)^{1/4})$ \cite{GratJeraToin22b}.

\subsection{Adding momentum (twice)}\label{sec:momentum}

\subsubsection{Using a momentum-based preconditioner}

Because of its widespread use, we now consider properties of variants
of \tal{\algname} augmented with momentum, and start with the
\tal{\algname.MM} given \vpageref{genalgomm}.

\algo{genalgomm}{\tal{\algname.MM}}{
  Given: a starting point $X_0$, operators $\{\calU_{k,\ell}\}_{k\ge 0,\ell\in\ii{L}}$ and
  $\{\calN_\ell\}_{\ell\in\ii{L}}$, and constants $\eta,\varsigma>0$ and
  $\mu_{\max}\in(0,1)$. 
  Set $\Gamma_{-1,\ell}=\varsigma I_\ell$ for $\ell\in\ii{L}$.\\*[1ex]
  For $k = 0, 1, \ldots$ \\
  \vspace*{-2mm}
  \begin{enumerate}
  \item Draw $\xi_k$ and compute $\tG_k = \tG(X_k,\xi_k)=
    (\tG_{k,1},\ldots,\tG_{k,L})$,
    \hspace*{9mm}\mbox{\small [gradient approx.]}
  \item On each block $\ell\in\ii{L}$, compute
    \begin{eqnarray}
    M_{k,\ell} & = &\!\!\!\left\{\begin{array}{ll}
      \tG_{0,\ell}                           &\tim{if} k = 0,\\
      \mu_k M_{k-1,\ell}+(1-\mu_k)\tG_{k,\ell} &\tim{if} k > 0,
      \end{array}\right.
    \hspace*{9.5mm}\mbox{\small [momentum M]}\label{Mmom}\\
    \Gamma_{k,\ell} & = &\Gamma_{k-1,\ell}+\calU_{k,\ell}(M_{k,\ell}),
    \hspace*{29mm}\mbox{\small [preconditioner update]} \label{Gamma-mom}  \\
    Z_{k,\ell}      & = & \Gamma_{k,\ell}^{-1/2}M_{k,\ell},
    \hspace*{41mm}\mbox{\small [preconditioned gradient]}\label{Zmom} \\
    X_{k+1,\ell}    & = & X_{k,\ell} -
    \eta\,\|Z_{k,\ell}\|_{*,\ell}\,\calN_\ell(Z_{k,\ell}).
    \hspace*{18mm}\mbox{\small [scaled normalized step]} \label{xkp1-mom}
    \end{eqnarray}
  \end{enumerate}
}

\noindent
This is \tal{\algname} where \req{z-def} and \req{Gamma-def} are replaced by
\req{Mmom}--\req{Zmom}.  This modification of the algorithm in
effect adds a specific type of momentum, although this is not the
only possible one.
Another variant where $\calU_{k,\ell}(\tG_{k,\ell})$ is used instead of
$\calU_{k,\ell}(M_{k,\ell})$ will be considered in Section~\ref{sec:mg}.

We now show that the theory of the previous section can be extended to cover
\tal{\algname.MM}. We start by establishing a bound on the norm of the
difference between the block-wise momentum $M_{k,\ell}$ and the
approximate gradient $\tG_{k,\ell}$.

\llem{lem:E-form}{
Suppose that Assumption~\ref{ass:smooth-op} holds. 
Define
\beqn{E-def}
E_{k,\ell} = M_{k,\ell}-\tG_{k,\ell} \tim{ for } k\ge 0 \tim{and} \ell\in\ii{L}.
\eeqn
Then
\beqn{errEsq}
\sum_{j=0}^k\E{\|E_j\|_*^2}
 \le
 \frac{6}{(1-\mu_{\max})^2}\sum_{j=0}^k\bar\mu_j^2\E{\|\tG_j-G_j\|_*^2}
 +\frac{3 L_G^2\eta^2}{(1-\mu_{\max})^2}\sum_{j=0}^k\bar\mu_j^2\E{\|Z_j\|_*^2},
 \eeqn
 where
 \beqn{barmu-def}
 \bar\mu_j= \max[\mu_j,\mu_{j+1}].
 \eeqn
}

\proof{If $k=0$, \req{errEsq} trivially holds.
  Suppose that $k\ge1$ and note that \req{E-def} gives that
\[
\begin{aligned}
  E_{k,\ell}
  & = \mu_kM_{k-1,\ell}+(1-\mu_k)\tG_{k,\ell} - \tG_{k,\ell}\\
  & = \mu_k(E_{k-1,\ell} + \tG_{k-1,\ell}) +(1-\mu_k)\tG_{k,\ell} - \tG_{k,\ell}\\
  & = \mu_k E_{k-1,\ell}+ \mu_k (\tG_{k-1,\ell}-\tG_{k,\ell}).
\end{aligned}
\]
Using
the facts that $\mu_k \le \mu_{\max}$, that $(a+b)^2\le (1+\tau)a^2
+ (1+1/\tau)b^2$ for any $\tau>0$ and that $(a+b+c)^2\le 3a^2 +
3b^2+3c^2$, we obtain that 
\[
\begin{aligned}
\|E_{k,\ell}\|_{*,\ell}^2
\le & (1+\tau)\mu_{\max}^2\|E_{k-1,\ell}\|_{*,\ell}^2\\
& + 3\mu_k^2\left(1+\frac{1}{\tau}\right)
\left(\|G_{k,\ell}-G_{k-1,\ell}\|_{*,\ell}^2+\|\tG_{k,\ell}-G_{k,\ell}\|_{*,\ell}^2+\|\tG_{k-1,\ell}-G_{k-1,\ell}\|_{*,\ell}^2\right)\\
\end{aligned}
\]
Summing over $\ell\in\ii{L}$ and using \req{dual-product-norm}, we obtain that
\[
\begin{aligned}
\|E_k\|_*^2
\le & (1+\tau)\mu_{\max}^2\|E_{k-1}\|_*^2\\
& + 3\mu_k^2\left(1+\frac{1}{\tau}\right)
\left(\|G_k-G_{k-1}\|_*^2+\sum_{\ell=1}^L\|\tG_{k,\ell}-G_{k,\ell}\|_{*,\ell}^2+\sum_{\ell=1}^L\|\tG_{k-1,\ell}-G_{k-1,\ell}\|_{*,\ell}^2\right)\\
\end{aligned}
\]
Observe now that Assumption~\ref{ass:smooth-op} with \req{xkp1-mom} and
\req{S-normed} imply that
\beqn{isLip}
\|G_k-G_{k-1}\|_*^2\le
L_G^2\sum_{\ell=1}^L\|X_{k,\ell}-X_{k-1,\ell}\|^2
= L_G^2\eta^2\sum_{\ell=1}^L\|Z_{k-1,\ell}\|_{*,\ell}^2\|\calN_\ell(Z_{k-1,\ell})\|_\ell^2 = L_G^2\eta^2 \|Z_{k-1}\|_*^2,
\eeqn
Now let $\tau = (1-\mu_{\max})/\mu_{\max}$. Then $(1+\tau)\mu_{\max}^2 = \mu_{\max} < 1$ and
$(1+1/\tau)\mu_k^2=\mu_k^2/(1-\mu_{\max})$. If
$\vartheta_k^2= \sum_{\ell=1}^L\E{\|\tG_{k,\ell}-G_{k,\ell}\|_{*,\ell}^2}=\E{\|\tG_k-G_k\|_*^2}$,
we have, after taking total expectation and using \req{isLip}
that
\[
\E{\|E_k\|_*^2}
\le \mu_{\max}\E{\|E_{k-1}\|_*^2}
+ \frac{3\mu_k^2}{1-\mu_{\max}}\left( L_G^2\eta^2 \E{\|Z_{k-1}\|_*^2} + \vartheta_k^2+\vartheta_{k-1}^2\right)
\]
By the definition of $\bar\mu_j$,
\[
\sum_{j=1}^k
\mu_j^2
\mathbb E\big[\|Z_{j-1}\|_*^2\big]
=
\sum_{j=0}^{k-1}
\mu_{j+1}^2
\mathbb E\big[\|Z_j\|_*^2\big]
\leq
\sum_{j=0}^k
\bar\mu_j^2
\mathbb E\big[\|Z_j\|_*^2\big].
\]
Moreover,
\[
\sum_{j=1}^k\mu_j^2\left(\vartheta_j^2+\vartheta_{j-1}^2\right)
=\sum_{j=1}^k \mu_j^2\vartheta_j^2
+\sum_{j=0}^{k-1}\mu_{j+1}^2\vartheta_j^2
\leq 2\sum_{j=0}^k \bar\mu_j^2\vartheta_j^2.
\]
Combining these inequalities therefore gives
\[
\sum_{j=0}^k
\mathbb E\big[\|E_j\|_*^2\big]
\leq
\frac{6}{(1-\mu_{\max})^2}
\sum_{j=0}^k
\bar\mu_j^2
\mathbb E\big[\|\widetilde G_j-G_j\|_*^2\big]
+
\frac{3L_G^2\eta^2}{(1-\mu_{\max})^2}
\sum_{j=0}^k
\bar\mu_j^2
\mathbb E\big[\|Z_j\|_*^2\big],
\]
which proves  \req{errEsq}.
}

\noindent
We now wish to apply the theory above by
considering that the momentum $M_{k,\ell}$ plays the role of the
approximate gradient (in Step~2 of \algname) in this theory.
In particular, this change of perspectives implies that
Assumption~\ref{ass:identities} now becomes
\begin{assumption}[Modified structural identities]\label{ass:mom-identities}
For all $k\ge 0$ and all $\ell \in \ii{L}$, we have that
\beqn{ineq1-mom}
\|Z_{k,\ell}\|_{*,\ell}\,\ip{M_{k,\ell},\calN_\ell(Z_{k,\ell})}_F
= \tr\!\big(\Gamma_{k,\ell}^{-1/2}\calU_{k,\ell}(M_{k,\ell})\big),
\eeqn
and
\beqn{ineq2-mom}
\|Z_{k,\ell}\|_{*,\ell}^2=\tr\!\big(\Gamma_{k,\ell}^{-1}\calU_{k,\ell}(M_{k,\ell})\big).
\eeqn
\end{assumption}
Fortunately, using our theory is possible because, as we show below,
\req{errEsq} implies that condition \req{var-cond} of 
Theorem~\ref{th:master-op} holds with suitable constants.  This allows
us to derive the analog of Theorem~\ref{thm:convergence} for the
``momentum'' variant of \algname.

\lthm{thm:conv-mom}{
Suppose that the \tal{\algname} algorithm modified by
\req{Mmom}--\req{Zmom} is applied to problem \req{problem} and that
Assumptions~\ref{ass:bounded-op}, \ref{ass:smooth-op}, \ref{ass:opttransfer} and
\ref{ass:mom-identities} hold. Then 
\beqn{therate-mom}
\frac{1}{k+1}\sum_{j=0}^k\E{\|G_j\|_*}
\le \frac{1}{\sqrt{k+1}}\Big((\kappa_\circ\,+1)\Theta_k
+\omega\sqrt{2N\log(\Theta_k)}+\omega\sqrt{\max[\kappa_d,1]}\Big)
\eeqn
where $\Theta_k$ is defined in \req{Thetak-def}--\req{kaps-def} using
\beqn{nuknown}
\nu_k^2 =
\left(\frac{12\,\mu_{\max}^2}{(1-\mu_{\max})^2}+2\right)\sum_{j=0}^k\E{\|\tG_j-G_j\|_*^2}
\tim{ and }
\omega^2 = \frac{6\,\mu_{\max}^2L_G^2\eta^2}{(1-\mu_{\max})^2}.
\eeqn
}

\proof{
Observe first that the identity $(a+b)^2\le 2a^2+2b^2$,
\req{errEsq} and the bound $\bar\mu_k\le\mu_{\max}$ give that
\beqn{MmG}
\begin{aligned}
 \sum_{j=0}^k\E{\|M_j-G_j\|_*^2}
&\le 2\sum_{j=0}^k\E{\|\tG_j-G_j\|_*^2}
     +2\sum_{j=0}^k\E{\|E_j\|_*^2}\\
&\le\left(\frac{12\,\mu_{\max}^2}{(1-\mu_{\max})^2}+2\right) \sum_{j=0}^k\E{\|\tG_j-G_j\|_*^2}
     +\frac{6\,\mu_{\max}^2L_G^2\eta^2}{(1-\mu_{\max})^2}\sum_{j=0}^k\E{\|Z_j\|_*^2}\\
\end{aligned}
\eeqn
which is \req{var-cond} with \req{nuknown}.
We may therefore apply Theorem~\ref{thm:convergence} (under
Assumption~\ref{ass:mom-identities}) with these values
and deduce that
\beqn{therate-M}
\sum_{j=0}^k \E{\|M_j\|_*} \le \sqrt{k+1}\,\kappa_\circ\,\Theta_k.
\eeqn
Hence, using the triangle, Cauchy-Schwarz and Jensen inequalities, we obtain that
\beqn{aaa}
\begin{aligned}
\sum_{j=0}^k\E{\|G_j\|_*}
&\leq \sum_{j=0}^k\E{\|M_j\|_*} + \sum_{j=0}^k\E{\|M_j-G_j\|_*}\\
&\leq \sum_{j=0}^k\E{\|M_j\|_*} + \sqrt{k+1} \sqrt{\sum_{j=0}^k \E{\|M_j-G_j\|_*}^2}\\
&\leq  \sqrt{k+1}\,\kappa_\circ\,\Theta_k + \sqrt{k+1} \sqrt{\sum_{j=0}^k \E{\|M_j-G_j\|_*^2}}\\
&\leq  \sqrt{k+1}\,\kappa_\circ\,\Theta_k + \sqrt{k+1} \sqrt{\nu_k^2 +\omega^2\sum_{j=0}^k\sum_{\ell=1}^L\E{\|Z_{j,\ell}\|_{*,\ell}^2}}\\
\end{aligned}
\eeqn
where we have inserted  \req{nuknown} in \req{MmG} to obtain the last equality.
But, successively using \req{Delta-low}, \req{Delta-upper}
and \req{Thetak-def}, we deduce that
\[
\sum_{j=0}^k\sum_{\ell=1}^L\E{\|Z_{j,\ell}\|_{*,\ell}^2}
\le \Delta_k
\le
\kappa_d+2N\log\left(\E{\sum_{\ell=1}^L\tr(\Gamma_{k,\ell}^{1/2})}\right)
\le \kappa_d+2N\log\left(\Theta_k\right),
\]
Substituting this inequality into \req{aaa} gives
\[
\begin{split}
\sum_{j=0}^k \mathbb{E}[\|G_j\|_*]
&\leq\sqrt{k+1}\,\kappa_\circ\Theta_k
+\sqrt{k+1}\sqrt{\nu_k^2+\omega^2\bigl(\kappa_d+2N\log(\Theta_k)\bigr)}\\
&\leq\sqrt{k+1}\left[\kappa_\circ\Theta_k+\nu_k+\omega\sqrt{\kappa_d+2N\log(\Theta_k)}\right].
\end{split}
\]
Since $\nu_k\leq\Theta_k$, we obtain
\[
\sum_{j=0}^k \mathbb{E}[\|G_j\|_*]
\leq\sqrt{k+1}\left[(\kappa_\circ+1)\Theta_k+\omega\sqrt{\kappa_d+2N\log(\Theta_k)}
\right].
\]
Dividing by $k+1$  and using $\sqrt{a+b}\le\sqrt{a}+\sqrt{b}$
therefore yields \req{therate-mom}.
}

\noindent
The difference between \req{thetruerate} and \req{therate-mom} is only
a constant and a logarithmic term in factor of $(k+1)^{-1/2}$.  As a
consequence, the global rate of convergence of the momentum variant of
\tal{\algname} is essentially identical to that of the version without
momentum, and Corollary~\ref{therate2} still applies to the momentum variant.

We finally note that our momentum definition and convergence analysis do not
make the assumption that the stepsize parameter $\eta$ is sufficiently
small, in contrast with previous proofs of convergence where results
assume that $\eta$ is small enough, in particular smaller than a multiple
of the (usually unknown) Lipschitz constant $L_G$
\cite{GuoYinJinYang21,HongLin24,XiaoHuLiuToh24}.

\subsubsection{Using a gradient based preconditioner}\label{sec:mg}

The momentum approach we have described above uses
\req{Mmom}--\req{Zmom}.  Another version can be considered, using
\req{Mmom} and \req{Zmom} but keeping the ``pure gradient'' technique
\req{Gamma-def} to accumulate the preconditioner, as shown \vpageref{genalgomg}.

\algo{genalgomg}{\tal{\algname.MG}}{
  Given: a starting point $X_0$ operators $\{\calU_{k,\ell}\}_{k\ge 0,\ell\in\ii{L}}$ and
  $\{\calN_\ell\}_{\ell\in\ii{L}}$,  and constants $\eta,\varsigma>0$ and $\mu_{\max}\in(0,1)$. Set
  $\Gamma_{-1,\ell}=\varsigma I_\ell$ for $\ell\in\ii{L}$.\\*[1ex]
  For $k = 0, 1, \ldots$ \\
  \vspace*{-2mm}
  \begin{enumerate}
  \item  Draw $\xi_k$ and compute $\tG_k =\tG(X_k,\xi_k) =
    (\tG_{k,1},\ldots,\tG_{k,L})$,
    \hspace*{8mm}\mbox{\small [gradient approx.]}
  \item On each block $\ell\in\ii{L}$, compute
    \begin{eqnarray}
    \Gamma_{k,\ell} & =
    &\Gamma_{k-1,\ell}+\calU_{k,\ell}(\tG_{k,\ell}),
    \hspace*{29mm}\mbox{\small [preconditioner update]} \label{Gammom2}  \\
    M_{k,\ell} & = &\!\!\!\left\{\begin{array}{ll}
      \tG_{0,\ell}                           &\tim{if} k = 0,\\
      \mu_k M_{k-1,\ell}+(1-\mu_k)\tG_{k,\ell} &\tim{if} k > 0,
      \end{array}\right.
    \hspace*{9.5mm}\mbox{\small [momentum G]}\label{Mmom2}\\
    Z_{k,\ell}      & = & \Gamma_{k,\ell}^{-1/2}M_{k,\ell},
    \hspace*{41mm}\mbox{\small [preconditioned gradient]}\label{Zmom2} \\
    X_{k+1,\ell}    & = & X_{k,\ell} -
    \eta\,\|Z_{k,\ell}\|_{*,\ell}\,\calN_\ell(Z_{k,\ell}).
    \hspace*{18mm}\mbox{\small [scaled normalized step]} \label{xkp1-mom2}
    \end{eqnarray}
  \end{enumerate}
  }

\noindent
As it turns out, it is also possible to derive a unified convergence
theory for this choice, albeit this requires stronger assumptions
and leads to a possibly worse rate of convergence.
This alternative theory hinges on the fact that
$\calU_{k,\ell}(\tG_{k,\ell})$ can be viewed as a perturbation of
$\calU_{k,\ell}(M_{k,\ell})$ , and therefore that the structural relations of
Assumption~\ref{ass:mom-identities} are themselves perturbed.
Fortunately, it remains possible to bound these perturbations by a mix
of variance-related and second-order terms, the first of which
potentially affecting the resulting global convergence
rate.  The final outcome is given by the following theorem and
corollary (whose detailed proofs can be found in the appendix).

\lthm{thm:convergence-alt}{
Suppose that the \tal{\algname} algorithm with \req{z-def} replaced by
\req{Mmom2} and \req{Zmom2} is applied to problem \req{problem}.
Suppose that Assumptions~\ref{ass:bounded-op}, \ref{ass:smooth-op}, \ref{ass:opttransfer} and
\ref{ass:mom-identities}  hold. If $\mu >0$, suppose also that
there exists constants $\kappa_\Box,\kappa_\diamond > 0$ such that
  \beqn{ass:sub-add}
  \calU_{k,\ell}(U+V) \preceq \kappa_\Box \calU_{k,\ell}(U) + \kappa_\Box \calU_{k,\ell}(V)
  \tim{ and }
  \tr\!\Big(\calU_{k,\ell}(U)\Big) \le \kappa_\diamond \|U\|_{*,\ell}^2
  \eeqn
  for all $\ell\in\ii{L}$, $k\ge0$ and all $U,V\in\Re^{n_\ell\times m_\ell}$,
  and that 
  \beqn{small-eta}
  \tim{either }
  \eta \le \frac{1-\mu_{\max}}{\mu_{\max}\, L_G}
  \sqrt{\frac{\varsigma}{6\kappa_\Box\kappa_\diamond}}
  \tim{ ~or~ }
  \sum_{j=0}^\infty \bar\mu_j^2\|Z_j\|_*^2  \le \kappa_{\mu Z}
  \eeqn
for some $\kappa_{\mu Z}\ge 0$ and $\bar\mu_j$ defined by \req{barmu-def}.
Then
\beqn{therate-alt}
\sum_{j=0}^k \E{\|\tG_j\|_*} \le \sqrt{k+1}\,\kappa_\circ\,\Theta_k
\eeqn
where
\beqn{Thetak-def-alt}
\Theta_k = \max\left[\,e^{\max\left[1,\frac{\kappa_d}{2N}\right]},
  \, 3(\kap{gap}+\varepsilon(\nu_k,\theta_k)), T_k,\,Y\,\right],
\eeqn
with $\nu_k$ defined in \req{var-cond},
\[
\varepsilon(\nu,\theta)
= \kappa_\Box\left(\sqrt{\kappa_{2,z}}\,\nu+\sqrt{\kappa_{2,\theta}}\,\nu\theta
+\kappa_{\theta\theta}\theta^2\right)
\]
\beqn{thetak-def}
\theta_k^2 = \sum_{j=0}^k\sum_{\ell=1}^L\bar\mu_j^2\E{\|\tG_{j,\ell}-G_{j,\ell}\|_{*,\ell}^2},
\eeqn
\[
T_k = 6(2\kappa_{\Box})^{3/2}\,\sqrt{N}
\nu_k\,\sqrt{\max\left[1,\log\left(6(2\kappa_{\Box})^{3/2}\sqrt{N}\,\,\nu_k\right)\right]},
\]
\[
Y =
24N\kappa_\Delta\kappa_\Box^2
\log\left(24N\kappa_\Delta\kappa_\Box^2\right),
\]
$\kappa_\circ$ being defined in Assumption~\ref{ass:opttransfer},
$\kappa_d$ in \req{kaps-def},
$\kappa_{2,z}$ and $\kappa_{2,\theta}$ in \req{kappa2} and
$\kap{gap}$, $\kappa_{\theta\theta}$ and
$\kappa_\Delta$ in \req{kaps-def1}--\req{kaps-def2}.
Moreover, if $\tG_k$ is an unbiased estimator of $G_k$,
i.e. $\Econd{k}{\tG_k}= G_k$ for all $k\ge 0$, then
\beqn{thetruerate-alt}
\min_{j\in\iiz{k}}\E{\|G_j\|_*}
\le \frac{1}{k+1}\sum_{j=0}^k \E{\|G_j\|_*}
\le \frac{\kappa_\circ\,\Theta_k}{\sqrt{k+1}}.
\eeqn
}

\noindent
As it turns out, assuming \req{ass:sub-add} is not restrictive in the
applications considered in this paper, as we briefly discuss below in
the appendix. Supposing \req{small-eta} is definitely less desirable because the
value of the Lipschitz constant $L_G$  or that of $\kappa_{\mu Z}$ (if
it exists) is usually unknown.  It is
however common practice in the literature
\cite{GuoYinJinYang21,HongLin24,XiaoHuLiuToh24}.
Note the term $\varepsilon(\nu_k,\theta_k)$ in the middle term of
\req{Thetak-def-alt}, which is the crucial difference between
\req{Thetak-def-alt} and \req{Thetak-def}.
Because $\theta_k < \nu_k$ (since $\mu_k\le \mu_{\max}<1$), the term
which achieves the fastest growth when $k$ tends to infinity in
\req{therate-alt} is the largest of $\nu_k\theta_k$ (the second term of
$\varepsilon(\nu_k,\theta_k)$) and $\nu_k\sqrt{\log(\nu_k)}$ (the
$T_k$ term), that is $\nu_k\max[\theta_k,\sqrt{\log(\nu_k)}]$. Thus
the rate of convergence of \tal{\algname.MG} is never better than that
of \tal{\algname} and is only as good when
$\theta_k=\calO\left(\sqrt{\log(\nu_k)}\right)$.
This induces the modified convergence rate expressed by the following
corollary.

\llcor{therate2-alt}{
Suppose that the \tal{\algname} algorithm with \req{z-def} replaced by
\req{Mmom2} and \req{Zmom2} is applied to problem \req{problem}.
Suppose that Assumptions~\ref{ass:bounded-op}, \ref{ass:smooth-op}, \ref{ass:opttransfer} and
\ref{ass:mom-identities} hold.
If $\mu>0$, suppose also that \req{ass:sub-add} and \req{small-eta}
hold, that
\beqn{betaisktogamma-alt}
\Econd{k}{\tG_k}= G_k
\tim{ and }
\Econd{k}{\|\tG_{k,\ell}-G_{k,\ell}\|_*^2} \le
\frac{\sigma_\ell^2}{ (k+1)^\alpha}
\eeqn
for some $\alpha>0$ and all $k \ge 0$ and $\ell\in\ii{L}$, and that
\[
\mu_k = \frac{\mu_{\max}}{(k+1)^\beta}
\]
for some $\mu_{\max}<1$ and some $\beta\ge 0$.
\beqn{detailed-rates-mg}
\frac{1}{k+1}\sum_{j=0}^k \E{\|G_j\|_*}=
\left\{\begin{array}{ll}
\calO\left(\bigfrac{1}{(k+1)^{\alpha+\beta-\half}}\right)
&\tim{if } \alpha+2\beta<1,\\
\calO\left(\bigfrac{\sqrt{\log(k+1)}}{(k+1)^{\sfrac{\alpha}{2}}}\right)
&\tim{if } \alpha<1 \tim{and} \alpha+2\beta\ge1,\\
\calO\left(\bigfrac{\log(k+1)}{\sqrt{k+1}}\right)
&\tim{if } \alpha = 1 \tim{and} \alpha+2\beta=1,\\*[2ex]
\calO\left(\bigfrac{\sqrt{\log(k+1)\log(\log(k+1))}}{\sqrt{k+1}}\right)
&\tim{if } \alpha = 1 \tim{and} \alpha+2\beta>1,\\*[2ex]
\calO\left(\bigfrac{1}{\sqrt{k+1}}\right)
&\tim{if } \alpha > 1.
\end{array}\right.
\eeqn
}

\noindent
Note that, when $\alpha+2\beta<1$,  the right-hand side of
  \req{detailed-rates-mg} tends to zero only if $\alpha+\beta> \half$.
One should however be careful not to identify better theoretical
convergence properties with better practical performance on specific
classes of problems. Indeed, limited numerical experiments
suggest that the second momentum variant might outperform the first.

\numsection{Application to existing algorithms}\label{sec:appl}

We now consider the four popular first-order algorithms introduced in
Sections~\ref{intro:adn} to \ref{intro:muon} and show that their
convergence behaviour is covered by the above results.
This is
achieved by verifying that the conditions \req{ineq1}, \req{ineq2} and
\req{eq:generic} hold in each case for all blocks concerned.
In all cases considered here, the preconditioner update
$\calU_{k,\ell}$ is independent of $k$. For simplicity of exposition,
we focus on the situation where there is no momentum ($\mu_k=0$). We
again use lower case symbols for vectors.

\subsection{AdaNorm \cite{StreMcMa10,DuchHazaSing11,WardWuBott19}}

With the choices specified by \req{adaN}, we have that
\[
\tr\left(\Gamma_{k,\ell}^{-1}\calU_{k,\ell}(\tG_{k,\ell}) \right)
= \tr\left(\frac{\|\tG_{k,\ell}\|_\ell^2}{n_\ell}\Gamma_{k,\ell}^{-1}\right).
\]
Because of the definition of $\Gamma_{k,\ell}$, this becomes
\[
\tr\left(\Gamma_{k,\ell}^{-1}\calU_{k,\ell}(\tG_{k,\ell}) \right)
= \frac{\|\tG_{k,\ell}\|_\ell^2}{\gamma_{k,\ell}}
= \|Z_{k,\ell}\|_\ell^2,
\]
which is \req{ineq2}. Similarly,
\[
\tr(\Gamma_{k,\ell}^{-1/2}\calU_{k,\ell}(\tG_{k,\ell}))
= \frac{\|\tG_{k,\ell}\|_\ell^2}{\sqrt{\gamma_{k,\ell}}}
= \|Z_{k,\ell}\|_\ell\,\ip{\tG_{k,\ell},\calN_\ell(Z_{k,\ell})},
\]
proving \req{ineq1}.
Moreover, Assumption~\ref{ass:opttransfer} also holds on block $\ell$.
Indeed, for any $\tG_\ell$, the second and last parts of \req{adaN} give
\[
\tr(\calU_{k,\ell}(\tG_\ell))=\|\tG_\ell\|_\ell^2=\|\tG_\ell\|_{*,\ell}^2,
\]
and thus \req{eq:generic} holds with $\kappa_\circ=1$.

Thus, whenever a block $\ell$ is endowed with this isotropic scalar
geometry, Theorem~\ref{thm:convergence} and Corollary~\ref{therate2}
apply to that block without modification.  The standard ``one block''
AdaNorm is obtained when there is only one block $(L=1, n_1=d_1=N)$.

\subsection{Full AdaGrad \cite{DuchHazaSing11}}

With the choices \req{adaF}, we have that
\[
\tr(\Gamma_{k,\ell}^{-1}\calU_{k,\ell}(\tG_{k,\ell}))
=\tr(\Gamma_{k,\ell}^{-1/2}\tG_{k,\ell}\tG_{k,\ell}^T\Gamma_{k,\ell}^{-1/2})
= \|Z_{k,\ell}\|_E^2,
\]
which is \req{ineq2}. Similarly,
\[
\tr(\Gamma_{k,\ell}^{-1/2}\calU_{k,\ell}(\tG_{k,\ell}))
=\tr(\Gamma_{k,\ell}^{-1/2}\tG_{k,\ell}\tG_{k,\ell}^T)
= \ip{\tG_{k,\ell},Z_{k,\ell}}
= \|Z_{k,\ell}\|_E\,\ip{\tG_{k,\ell},\calN_\ell(Z_{k,\ell})},
\]
proving \req{ineq1}.
Moreover, Assumption~\ref{ass:opttransfer} also holds on block $\ell$.
Indeed, for any $G_\ell$, the second and last part of \req{adaF} give
\[
\tr(\calU_{k,\ell}(\tG_\ell))
=
\tr(\tG_\ell \tG_\ell^T)
=
\|\tG_\ell\|_E^2
=\|\tG_\ell\|_{*,\ell}^2
\]
so that \req{eq:generic} holds with $\kappa_\circ=1$.

Thus, whenever a block $\ell$ is endowed with this full matrix geometry,
Theorem~\ref{thm:convergence}, and Corollary~\ref{therate2} apply to that block without modification.

\subsection{Diagonal AdaGrad \cite{DuchHazaSing11,McMaStre10}}

With the choices \req{adaD}, we have that
\[
\tr(\Gamma_{k,\ell}^{-1}\calU_{k,\ell}(\tG_{k,\ell}))
= \sum_{i=1}^{n_\ell} \frac{[\tG_{k,\ell}]_i^2}{[\Gamma_{k,\ell}]_{ii}}
= \|Z_{k,\ell}\|_E^2,
\]
which proves \req{ineq2}. Moreover,
\[
\tr(\Gamma_{k,\ell}^{-1/2}\calU_{k,\ell}(\tG_{k,\ell}))
=
\sum_{i=1}^{n_\ell} \frac{[\tG_{k,\ell}]_i^2}{\sqrt{[\Gamma_{k,\ell}]_{ii}}}
=
\ip{\tG_{k,\ell},Z_{k,\ell}}
=
\|Z_{k,\ell}\|_E\,\ip{\tG_{k,\ell},\calN_\ell(Z_{k,\ell})},
\]
which proves \req{ineq1}.
Moreover, Assumption~\ref{ass:opttransfer} also holds in this case.
Indeed, for any $G_\ell \in \mathbb{R}^{n_\ell}$,
\[
\calU_{k,\ell}(G_\ell)=\diag\big([G_\ell]_i^2\big),
\]
and therefore
\[
\tr(\calU_{k,\ell}(G_\ell))
= \sum_{i=1}^{n_\ell} [G_\ell]_i^2
= \|G_\ell\|_E^2
= \|G_\ell\|_{*,\ell}^2,
\]
so that \req{eq:generic} holds with $\kappa_\circ=1$.

Thus, as above, Theorem~\ref{thm:convergence} and Corollary~\ref{therate2}
apply without modification to this diagonal AdaGrad geometry on block $\ell$.

\subsection{Adaptive MUON/AdaGO \cite{Jordetal24,SiZhanShen25,ZhanLiuScha25}}

Consider the choices \req{MUON-N} and \req{MUON-choices}.
For $\ell \in \calA$, the verification of \req{ineq1}, \req{ineq2}
and \req{eq:generic} is identical to that of AdaNorm.
For $\ell\in\calM$, we have
\[
\tr(\Gamma_{k,\ell}^{-1}\calU_{k,\ell}(\tG_{k,\ell}))
= \frac{\|\tG_{k,\ell}\|_*^2}{\gamma_{k,\ell}}
= \|Z_{k,\ell}\|_*^2,
\]
which proves \req{ineq2}.
Therefore,
\[
\|Z_{k,\ell}\|_*\,\ip{\tG_{k,\ell},\calN_\ell(Z_{k,\ell})}
= \frac{\|\tG_{k,\ell}\|_*^2}{\sqrt{\gamma_{k,\ell}}}
= \tr(\Gamma_{k,\ell}^{-1/2}\calU_{k,\ell}(\tG_{k,\ell})),
\]
which proves \req{ineq1}.
 Moreover, Assumption~\ref{ass:opttransfer} also holds in this case.
Indeed, for any block $\ell$ and any $\tG_\ell$,
\[
\tr(\calU_{k,\ell}(\tG_\ell))=\left\{\begin{array}{ll}
\tr\left(\frac{\|\tG_\ell\|_*^2}{d_\ell} I_{d_\ell}\right) = \|\tG_\ell\|_*^2 & \tim{if } \ell\in\calM,\\[1mm]
\tr\left(\frac{\|\tG_\ell\|_E^2}{d_\ell} I_{d_\ell}\right) =
\|\tG_\ell\|_E^2 = \|\tG_\ell\|_*^2 & \tim{if } \ell\in\calA.
\end{array}\right.
\]
Thus \req{eq:generic} holds with $\kappa_\circ=1$ for all $\ell\in\ii{L}$.
Therefore Theorem~\ref{thm:convergence} and Corollary~\ref{therate2}
apply without modification to this adaptive MUON geometry.

\numsection{Conclusions}\label{sec:conclusions}

We have provided a general framework for first-order optimization
algorithms using adaptively preconditioned gradients but no function values. We
have shown that this framework covers a few of the popular adaptive
first-order methods, including AdaGrad, full AdaGrad, AdaNorm and
adaptive Muon, as well as any Cartesian mix of these. We have provided a fully stochastic
rate-of-convergence analysis for all methods in the
framework, with <and without momentum, using reasonable (although
necessarily stonger than affine variance)  assumptions
on the variance of the gradient oracle and without assuming bounded
stochastic gradients or small enough stepsizes.

Our results open several theoretical research issues. We anticipate
that, as is the case for AdaGrad, the framework can be adapted to
handle bounds \cite{BellGratMoriToin25} or more general convex
\cite{GuptKoreSing18,DaviDrus19,AlacLyu23} or nonconvex
\cite{FangNaMahoKola24,BellGratMoriToin26,GratToin26b,WangPierZhouCurt26}
constraints, as well as second-order information (should it be
available), but this requires verifications that are beyond the scope
of the present paper. Other questions also remain. Can the theory be
extended to cover other methods, like RMSProp \cite{TielHint12}
{or Shampoo \cite{GuptKoreSing18}? Is approximate
normalization (i.e. approximate $\calN_\ell(\cdot)$) acceptable? Can the
generality of the preconditioner update be exploited to derive new
preconditioners? Does the Cartesian framework allow more
general algorithmic settings?
Can the freedom to choose a different preconditioner update at
each iteration be used to advantage?
All these topics are the subject of ongoing
investigation.

{\footnotesize

\section*{\footnotesize Acknowledgments}

The authors are much indebted to two committed and timely referees
for their patience with an imperfect manuscript and their many helpful
and perceptive suggestions.  Philippe Toint is also grateful for the continued
friendly support of the APO team at ENSEEIHT (Toulouse).

\section*{\footnotesize Conflict of Interest}

The authors confirm that there is no conflict of interest related to
this paper.


}

\appendix

\appnumsection{Proofs for Theorem~\ref{thm:convergence-alt} and Corollary~\ref{therate2-alt}}

The definition \req{Gamma-def} immediately implies the following
bounds.

\llem{lem:Gpreceq}{
We have that, for all $k\ge 0$ and all $\ell\in\ii{L}$,
$\Gamma_{k,\ell}$ is symmetric positive definite and 
\beqn{Gamposdef}
\Gamma_{k,\ell}^{-1/2}\preceq \frac{1}{\sqrt{\varsigma}}I_\ell
\tim{and}
\Gamma_{k,\ell}^{-1}\preceq \frac{1}{\varsigma}I_\ell
\eeqn
}
\proof{Directly result from
  \req{Gamma-def}, the definition of $\calU_{k,\ell}$ and $\Gamma_{k,\ell} 
= \Gamma_{-1,\ell}+ \sum_{j=0}^k \calU_{j,\ell}(\tG_{j,\ell})
\succeq \Gamma_{-1,\ell} = \varsigma I_\ell$.
}

We now consider the impact of using momentum (i.e. $\mu>0$ and
$Z_{k,\ell}$ is a multiple of $M_{k,\ell}$) on the
structural identities of Assumption~\ref{ass:mom-identities}.

\llem{lem:perturbs}{
Suppose that $\mu>0$ and that Assumptions~\ref{ass:smooth-op} and \ref{ass:mom-identities}
and \req{ass:sub-add} hold.
Then
\beqn{pineq1}
\!\|Z_{k,\ell}\|_{*,\ell}\ip{\tG_{k,\ell},\calN_\ell(Z_{k,\ell})}_F
\!\ge \!\frac{1}{\kappa_\Box}\!\left[\tr\!\big(\Gamma_{k,\ell}^{-1/2}\calU_{k,\ell}(\tG_{k,\ell})\big)
\!-\!\tr\!\big(\Gamma_{k,\ell}^{-1/2}\calU_{k,\ell}(E_{k,\ell})\big)\right]
\!-\!\|Z_{k,\ell}\|_{*,\ell}\|E_{k,\ell}\|_{*,\ell}
\eeqn
and
\beqn{pineq2}
\|Z_{k,\ell}\|_{*,\ell}^2
\le \kappa_\Box \!\tr\!\big(\Gamma_{k,\ell}^{-1}\calU_{k,\ell}(\tG_{k,\ell})\big)
    + \kappa_\Box \tr\!\big(\Gamma_{k,\ell}^{-1}\calU_{k,\ell}(E_{k,\ell})\big)
\eeqn
}

\proof{
Using \req{E-def}, the linearity of the inner product and \req{ineq1-mom}, we have that
\beqn{S3.28}
\begin{aligned}
\|Z_{k,\ell}\|_{*,\ell}\,\ip{\tG_{k,\ell},\calN_\ell(Z_{k,\ell})}_F
&= \|Z_{k,\ell}\|_{*,\ell}\,\ip{M_{k,\ell},\calN_\ell(Z_{k,\ell})}_F
  - \|Z_{k,\ell}\|_{*,\ell}\,\ip{E_{k,\ell},\calN_\ell(Z_{k,\ell})}_F\\
&= \tr\!\big(\Gamma_{k,\ell}^{-1/2}\calU_{k,\ell}(M_{k,\ell})\big)
  - \|Z_{k,\ell}\|_{*,\ell}\,\ip{E_{k,\ell},\calN_\ell(Z_{k,\ell})}_F.
\end{aligned}
\eeqn
Now observe that \req{ass:sub-add} and \req{E-def} give that
\[
\calU_{k,\ell}(\tG_{k,\ell})
= \calU_{k,\ell}(M_{k,\ell}-E_{k,\ell})
\preceq \kappa_\Box \calU_{k,\ell}(M_{k,\ell}) + \kappa_\Box \calU_{k,\ell}(E_{k,\ell})
\]
Because $\Gamma_{k,\ell}^{-1/2}$ is positive definite
(Lemma~\ref{lem:Gpreceq}) and the
trace is monotone for the L\"{o}wner semi-order, we have that
\[
\tr\!\big(\Gamma_{k,\ell}^{-1/2}\calU_{k,\ell}(\tG_{k,\ell})\big)
\le \kappa_\Box\tr\!\big(\Gamma_{k,\ell}^{-1/2} \calU_{k,\ell}(M_{k,\ell})^22\big) + \kappa_\Box
\tr\!\big(\Gamma_{k,\ell}^{-1/2} \calU_{k,\ell}(E_{k,\ell})\big).
\]
Substituting the resulting lower bound on
$\Gamma_{k,\ell}^{-1/2}\calU_{k,\ell}(M_{k,\ell})$ into \req{S3.28}, we
deduce that
\[
\begin{aligned}
\|Z_{k,\ell}\|_{*,\ell}\,\ip{\tG_{k,\ell},\calN_\ell(Z_{k,\ell})}_F
&\ge \frac{1}{\kappa_\Box}\left[\tr\!\big(\Gamma_{k,\ell}^{-1/2}\calU_{k,\ell}(\tG_{k,\ell})\big)
  -\tr\!\big(\Gamma_{k,\ell}^{-1/2}\calU_{k,\ell}(E_{k,\ell})\big)\right]\\
&\hspace*{10mm} -\|Z_{k,\ell}\|_{*,\ell}\,\ip{E_{k,\ell},\calN_\ell(Z_{k,\ell})}_F\\
\end{aligned}
\]
and \req{pineq1} then results from the fact that, by duality of the primal and dual norms
and \req{S-normed},
\[
\ip{E_{k,\ell},\calN_\ell(Z_{k,\ell})}_F
\le \|E_{k,\ell}\|_{*,\ell}\|\calN_\ell(Z_{k,\ell})\|_\ell
= \|E_{k,\ell}\|_{*,\ell}.
\]
In order to prove \req{pineq2}, we first note that \req{E-def} and
\req{ass:sub-add} yield that
\[
\calU_{k,\ell}(M_{k,\ell})
= \calU_{k,\ell}(\tG_{k,\ell}+E_{k,\ell})
\preceq \kappa_\Box \calU_{k,\ell}(\tG_{k,\ell}) + \kappa_\Box \calU_{k,\ell}(E_{k,\ell})
\]
and thus, again using the monotonicity of the trace for the L\"{o}wner semi-order,
that
\[
\tr\!\big(\Gamma_{k,\ell}^{-1}\calU_{k,\ell}(M_{k,\ell})\big)
\le \kappa_\Box\tr\!\big(\Gamma_{k,\ell}^{-1}\calU_{k,\ell}(\tG_{k,\ell})\big)
+ \kappa_\Box\tr\!\big(\Gamma_{k,\ell}^{-1}\calU_{k,\ell}(E_{k,\ell})\big).
\]
Substituting this inequality in \req{ineq2-mom} then gives \req{pineq2}.
}

\noindent
Observe that \req{pineq1} is (up to a constant factor) a perturbation of a version of
\req{ineq1-mom} for block $\ell$ at iteration $k$, the perturbation term being
$\tr\!\big(\Gamma_{k,\ell}^{-1/2}\calU_{k,\ell}(E_{k,\ell})\big)
+\|Z_{k,\ell}\|_{*,\ell}\|E_{k,\ell}\|_{*\ell}$.
The same is true for \req{pineq2}, in which case the perturbation of
\req{ineq2-mom} is $\tr\!\big(\Gamma_{k,\ell}^{-1}\calU_{k,\ell}(E_{k,\ell})\big)$.
We now show that the terms occurring in these perturbations can be bounded.

\llem{lem:pert-bounds}{
  Suppose that Assumption~\ref{ass:smooth-op} and \ref{ass:mom-identities}
  and \req{ass:sub-add} hold.  Then
  \beqn{prod-bound}
  \sum_{j=0}^k\sum_{\ell=1}^L\E{\|Z_{j,\ell}\|_{*,\ell}\|E_{j,\ell}\|_{*,\ell}}
  \le \frac{3\theta_k^2}{(1-\mu_{\max})^2}
    +\frac{3 L_G^2\eta^2}{2(1-\mu_{\max})^2}\sum_{j=0}^k\bar\mu_j^2\E{\|Z_j\|_*^2}    
    + 2 \sum_{j=0}^k\E{\|Z_j\|_*^2}
  \eeqn
  \beqn{pert1-bound}
  \begin{aligned}
  \sum_{j=0}^k\sum_{\ell=1}^L\E{\tr\!\big(\Gamma_{j,\ell}^{-1/2}\calU_{k,\ell}(E_{j,\ell})\big)}
  &\le \frac{6\kappa_\diamond\theta_k^2}{(1-\mu_{\max})^2\sqrt{\varsigma}}
     + \frac{3\kappa_\diamond L_G^2\eta^2}{(1-\mu_{\max})^2\sqrt{\varsigma}}\,
       \sum_{j=0}^k\bar\mu_j^2\E{\|Z_j\|_*^2}
   \end{aligned}
  \eeqn
  \vspace*{-2mm}
  and
  \beqn{pert2-bound}
  \sum_{j=0}^k\sum_{\ell=1}^L\E{\tr\!\big(\Gamma_{k,\ell}^{-1}\calU_{k,\ell}(E_{k,\ell})\big)}
  \le \frac{6\kappa_\diamond\theta_k^2}{(1-\mu_{\max})^2\varsigma}
       + \frac{3\kappa_\diamond L_G^2\eta^2}{(1-\mu_{\max})^2\varsigma}\,
       \sum_{j=0}^k\bar\mu_j^2\E{\|Z_j\|_*^2}.
  \eeqn
}
  
\proof{
For each block $\ell\in\ii{L}$ and each iteration $j\in\iiz{k}$, we
have, using $ab\le (a^2+b^2)/2$, that
\[
\|Z_{k,\ell}\|_{*,\ell}\|E_{k,\ell}\|_{*,\ell}
\le \frac{1}{2}\|Z_{k,\ell}\|_{*,\ell}^2+ \frac{1}{2}\|E_{k,\ell}\|_{*,\ell}^2
\]
and thus
\[
\sum_{j=0}^k\sum_{\ell=1}^L\E{\|Z_{j,\ell}\|_{*,\ell}\|E_{j,\ell}\|_{*,\ell}}
\le \frac{1}{2} \sum_{j=0}^k\sum_{\ell=1}^L\E{\|Z_{j,\ell}\|_{*,\ell}^2}
+ \frac{1}{2}\sum_{j=0}^k\sum_{\ell=1}^L\E{\|E_{j,\ell}\|_{*,\ell}^2}
\]
Substituting now \req{errEsq} in this inequality gives \req{prod-bound}.
Consider now the trace terms.  Because of \req{Gamposdef}, the
monotonicity of the trace for the L\"{o}wner semi-order and
Assumption~\ref{ass:sub-add}, we have that 
\[
\tr\!\big(\Gamma_{k,\ell}^{-1/2}\calU_{k,\ell}(E_{k,\ell})\big)
\le \frac{1}{\sqrt{\varsigma}}\tr\!\big(\calU_{k,\ell}(E_{k,\ell})\big)
\le \frac{\kappa_\diamond}{\sqrt{\varsigma}} \|E_{k,\ell}\|_{*,\ell}^2
\]
Taking full expectation, summing for $j\in\iiz{k}$ and $\ell\in\ii{L}$
and substituting \req{errEsq} then gives \req{pert1-bound}.
The proof of \req{pert2-bound} is entirely similar, using $\varsigma$
instead of $\sqrt{\varsigma}$.
} 

\noindent
We may now substitute the bounds of Lemma~\ref{lem:pert-bounds} into those
of Lemma~\ref{lem:perturbs}.

\llem{lem:pert-bounds2}{
Suppose that $\mu>0$ and that Assumption~\ref{ass:smooth-op} and \ref{ass:mom-identities} and
\req{ass:sub-add} hold. Suppose in addition that \req{small-eta} is satisfied.
Then
\beqn{bound-p1}
\begin{aligned}
\bigsum_{j=0}^k\bigsum_{\ell=1}^L\E{\|Z_{j,\ell}\|_{*,\ell}\,\ip{\tG_{j,\ell},\calN_\ell(Z_{j,\ell})}_F}
\ge &\bigfrac{1}{\kappa_\Box}\bigsum_{j=0}^k\bigsum_{\ell=1}^L\E{\tr\!\big(\Gamma_{j,\ell}^{-1/2}\calU_{j,\ell}(\tG_{k,\ell})\big)}\\
&-\kappa_{1,\theta}\,\theta_k^2-
\kappa_{1,z}\,\bigsum_{j=0}^k\E{\|Z_j\|_*^2}
\end{aligned}
\eeqn
\vspace*{-3mm}%
with
\beqn{kappa1}
\kappa_{1,\theta} = \frac{6\kappa_\diamond}{(1-\mu_{\max})^2\sqrt{\varsigma}}
+ \frac{3}{(1-\mu_{\max})^2}
\tim{and}
\kappa_{1,z} =
\frac{3\kappa_\diamond \mu_{\max}^2L_G^2\eta^2}{(1-\mu_{\max})^2\sqrt{\varsigma}}
+\frac{3\mu_{\max}^2L_G^2\eta^2}{2(1-\mu_{\max})^2}+2,
\eeqn
and
\beqn{bound-p2}
\sum_{j=0}^k\E{\|Z_j\|_*^2}
\le 2 \kappa_\Box\sum_{j=0}^k\sum_{\ell=1}^L\E{\tr\!\big(\Gamma_{j,\ell}^{-1}\calU_{j,\ell}(\tG_{j,\ell})\big)}
    + \kappa_{2,\theta}\,\theta_k^2+ \kappa_{2,z},
\eeqn
\vspace*{-3mm}%
where
\beqn{kappa2}
\kappa_{2,\theta} =
\frac{12\kappa_\Box\kappa_\diamond}{(1-\mu_{\max})^2\varsigma}
\tim{ and }
\kappa_{2,z} = \left\{\begin{array}{ll}
0
&\tim{if the first part of \req{small-eta} holds,}\\
\frac{3\kappa_\Box\kappa_\diamond
  L_G^2\eta^2}{(1-\mu_{\max})^2\varsigma}\,\kappa_{\mu Z} 
&\tim{if the second part of \req{small-eta} holds.}
\end{array}\right.
\eeqn
}
\proof{
The inequality \req{bound-p1} is obtained by substituting
\req{prod-bound} and \req{pert1-bound} into \req{pineq1},
Moreover, taking the expectation in \req{pineq2}, summing for
$j\in\iiz{k}$ and $\ell\in\ii{L}$ and substituting \req{pert2-bound} gives that
\[
\begin{aligned}
\sum_{j=0}^k\sum_{\ell=1}^L \E{\|Z_{k,\ell}\|_{*,\ell}^2}
&\le \kappa_\Box\sum_{j=0}^k\sum_{\ell=1}^L \E{\tr\!\big(\Gamma_{k,\ell}^{-1}\calU_{k,\ell}(\tG_{j,\ell})\big)}
    +  \frac{6\kappa_\Box\kappa_\diamond}{(1-\mu_{\max})^2\varsigma}\,\theta_k^2\\
& \hspace*{2cm}      + \frac{3\kappa_\Box\kappa_\diamond L_G^2\eta^2}{(1-\mu_{\max})^2\varsigma}\,
       \sum_{j=0}^k\bar\mu_j^2\E{\|Z_j\|_*^2}.
\end{aligned}
\]
Suppose first that the first part of \req{small-eta} holds. Then,
since $\bar\mu_j\le \mu_{\max}$,    
\[
\frac{3\kappa_\Box\kappa_\diamond \mu_{\max}^2L_G^2\eta^2}{(1-\mu_{\max})^2\varsigma} \le \frac{1}{2}
\]
and \req{bound-p2} follows with
\[
\kappa_{2,z}=0
\tim{and}
\kappa_{2,\theta} = \frac{12\kappa_\Box\kappa_\diamond}{(1-\mu_{\max})^2\varsigma}.
\]
Alternatively, if
the second part of \req{small-eta} holds, then \req{bound-p2} follows
with
\[
\kappa_{2,z}=\frac{3\kappa_\Box\kappa_\diamond
  L_G^2\eta^2}{(1-\mu_{\max})^2\varsigma}\,\kappa_{\mu Z}
\tim{and}
\kappa_{2,\theta} = \frac{6\kappa_\Box\kappa_\diamond}{(1-\mu_{\max})^2\varsigma}.
\]
}

\noindent
From this point on, the analysis follows the lines of the argument of
Section~\ref{sec:theory} with Lemma~\ref{lem:pert-bounds2} providing an alternate set of
structural inequalities. Lemmas~\ref{sqrt-trace} to \ref{lem:sqrt-op}
are unchanged. The modifications to Theorem~\ref{th:master-op} are
minor.  It is restated as follows.

\lthm{th:master-op-mom}{
Suppose that Assumptions~\ref{ass:bounded-op}, \ref{ass:smooth-op}
and \ref{ass:mom-identities}, \req{var-cond} and \req{small-eta} hold.
Define $\Delta_k$ as in \req{Delta-def}. Then, for every $k\geq0$,
\beqn{master-mom}
\E{\sum_{\ell=1}^L\tr(\Gamma_{k,\ell}^{1/2})}
\le \kap{gap}+\varepsilon(\nu_k,\theta_k)
+\sqrt{2\kappa_\Box}\,\kappa_\Box\, \nu_k\sqrt{\Delta_k}+\kappa_\Delta\kappa_\Box^2\Delta_k,
\eeqn
where
\beqn{kaps-def1}
\kap{gap}
\eqdef \kappa_\Box\, \frac{f(X_0)-f_{\rm low}}{\eta}+\sqrt{\varsigma}\,N
+\kappa_\Box\left(\kappa_{1,z} +\omega+\frac{L_G\eta}{2}\right)\kappa_{2,z},
\eeqn
\beqn{vareps-def}
\varepsilon(\nu,\theta)
= \kappa_\Box\left(\sqrt{\kappa_{2,z}}\,\nu+\sqrt{\kappa_{2,\theta}}\,\nu\theta
+\kappa_{\theta\theta}\theta^2\right)
\eeqn
and
\beqn{kaps-def2}
\kappa_{\theta\theta}
\eqdef \kappa_{1,\theta}+\left(\kappa_{1,z}+\omega+\frac{L_G\eta}{2}\right)\kappa_{2,\theta},
\ms
\kappa_\Delta
\eqdef 2\left(\kappa_{1,z}+\omega+\frac{L_G\eta}{2}\right).
\eeqn
}

\proof{
Inequality~\req{bound-p1} gives, in place of \req{lin-term},
\[
\eta \sum_{j=0}^k\sum_{\ell=1}^L\E{\|Z_{j,\ell}\|_{*,\ell}\ip{\tG_{j,\ell}}{\calN_\ell(Z_{j,\ell})}}
\ge
\frac{\eta}{\kappa_\Box}\E{\sum_{\ell=1}^L\tr(\Gamma_{k,\ell}^{1/2})
-\sum_{\ell=1}^L\tr(\Gamma_{-1,\ell}^{1/2})}-\eta\kappa_{1,\theta}\theta_k^2-\eta\kappa_{1,z}\zeta_k,
\]
where
\[
\zeta_k
\eqdef
\sum_{j=0}^k\sum_{\ell=1}^L
\E{\|Z_{j,\ell}\|_{*,\ell}^2}.
\]

Proceeding as in the proof of Theorem~\ref{th:master-op}, while
keeping the stochastic error $\tG_{j,\ell}-G_{j,\ell}$ separate,
we obtain
\[
\begin{aligned}
&\frac{\eta}{\kappa_\Box}
\E{\sum_{\ell=1}^L\tr(\Gamma_{k,\ell}^{1/2})-\sum_{\ell=1}^L\tr(\Gamma_{-1,\ell}^{1/2})}
\\
&\quad\le
f(X_0)-f_{\rm low}+\eta\kappa_{1,\theta}\theta_k^2
+\eta\sum_{j=0}^k\sum_{\ell=1}^L\E{\|Z_{j,\ell}\|_{*,\ell}\|\tG_{j,\ell}-G_{j,\ell}\|_{*,\ell}}
+\left(\eta\kappa_{1,z}+\frac{L_G\eta^2}{2}\right)\zeta_k.
\end{aligned}
\]
By the Cauchy--Schwarz inequality and \req{var-cond},
\[
\begin{aligned}
\sum_{j=0}^k\sum_{\ell=1}^L\E{\|Z_{j,\ell}\|_{*,\ell}\|\tG_{j,\ell}-G_{j,\ell}\|_{*,\ell}}
&\le\sqrt{\sum_{j=0}^k\sum_{\ell=1}^L\E{\|Z_{j,\ell}\|_{*,\ell}^2}}
\sqrt{\sum_{j=0}^k\sum_{\ell=1}^L\E{\|\tG_{j,\ell}-G_{j,\ell}\|_{*,\ell}^2}}\\
&\le\sqrt{\zeta_k}\,\sqrt{\nu_k^2+\omega^2\zeta_k}\\
&\le\nu_k\sqrt{\zeta_k}+\omega\zeta_k.
\end{aligned}
\]
Hence
\[
\frac{\eta}{\kappa_\Box} \E{\sum_{\ell=1}^L\tr(\Gamma_{k,\ell}^{1/2})
-\sum_{\ell=1}^L\tr(\Gamma_{-1,\ell}^{1/2})}
\!\le\! f(X_0)-f_{\rm low}+\eta\kappa_{1,\theta}\theta_k^2
+\eta\nu_k\sqrt{\zeta_k}+\eta\!\left(\kappa_{1,z}+\omega+\frac{L_G\eta}{2}\right)\!\zeta_k.
\]
On the other hand, the definition of $\zeta_k$, \req{bound-p2},
\req{log-pot} and \req{Delta-def} give
\[
\zeta_k
\le\kappa_{2,z}+\kappa_{2,\theta}\theta_k^2
+2\kappa_\Box\sum_{j=0}^k\sum_{\ell=1}^L\E{\tr\!\left(\Gamma_{j,\ell}^{-1}\calU_{j,\ell}(\tG_{j,\ell})\right)}
\le\kappa_{2,z}+\kappa_{2,\theta}\theta_k^2+2\kappa_\Box\Delta_k.
\]
Consequently, using
$\sqrt{a+b+c}\leq\sqrt a+\sqrt b+\sqrt c$,
\[
\sqrt{\zeta_k}
\leq \sqrt{\kappa_{2,z}}+ \sqrt{\kappa_{2,\theta}}\,\theta_k+\sqrt{2\kappa_\Box\Delta_k}.
\]
Substitution in the preceding inequality yields
\[
\begin{aligned}
\frac{\eta}{\kappa_\Box}
\E{
\sum_{\ell=1}^L\tr(\Gamma_{k,\ell}^{1/2})-\sum_{\ell=1}^L\tr(\Gamma_{-1,\ell}^{1/2})}
&\le
f(X_0)-f_{\rm low}+\eta\nu_k\sqrt{\kappa_{2,z}}+\eta\nu_k\sqrt{\kappa_{2,\theta}}\,\theta_k\\
&\ms +\eta\left[\kappa_{1,\theta}+\left(\kappa_{1,z}
  +\omega+\frac{L_G\eta}{2}\right)\kappa_{2,\theta}\right]\theta_k^2\\
&\ms + \sqrt{2\kappa_\Box}\eta\nu_k\sqrt{\Delta_k}
     +2\eta \kappa_\Box\left(\kappa_{1,z}+\omega+\frac{L_G\eta}{2}\right)\Delta_k\\
&\ms
+\eta\left(\kappa_{1,z}+\omega+\frac{L_G\eta}{2}\right)\kappa_{2,z}.
\end{aligned}
\]
Finally, since
$\Gamma_{-1,\ell}=\varsigma I_\ell$ for all $\ell\in\ii{L}$,
\[
\sum_{\ell=1}^L\tr(\Gamma_{-1,\ell}^{1/2})=\sqrt{\varsigma}\,N,
\]
and the result follows from the definitions \req{kaps-def1}--\req{kaps-def2}.
}

\noindent
Note that the form of bound \req{master-mom} differs very little from that
of \req{master}: besides using different constants, \req{master-mom}
now involves the term $\varepsilon(\nu_k,\theta_k)$.

Lemmas~\ref{lem:spectral-log-square} and \ref{lem:delta-closure} are
again unmodified. Because of the similarity between \req{master-mom}
and \req{master}, the proof of Theorem~\ref{thm:convergence-alt} is
then identical to those of Theorems~\ref{thm:convergence-theta} and
\ref{thm:convergence}, except that the crucial inequality \req{ineqtk}
in Theorem~\ref{thm:convergence-theta} now holds with \req{abc-def}
replaced by 
\vspace*{-1mm}%
\[
a = \kap{gap}+\varepsilon(\nu_k,\theta_k),
\ms
b = 2\kappa_\Box\sqrt{2\kappa_\Box}\,\sqrt{N}
\tim{ and }
c =4\kappa_\Delta\kappa_\Box^2N.
\]

Finally, the proof of Corollary~\ref{therate2-alt} is similar to that
of Corollary~\ref{therate2}.
When considering \tal{\algname.MG} rather than \tal{\algname}, we see
from \req{therate-alt} that now
\beqn{Theta-mg}
\Theta_k = \calO\left(\nu_k\max\left[\theta_k,\sqrt{\log(\nu_k)}\right]\right)
\eeqn
instead of \req{Theta-nomg}, and we may derive (as done in Corollary~\ref{therate2} for \req{nuk-vals}) that
\beqn{thetak-vals}
\theta_k \le \left\{\begin{array}{ll}
\bigfrac{\sigma_{\rm tot}\mu_{\max}}{\sqrt{1-\alpha-2\beta}}(k+1)^{\half(1-\alpha)-\beta}
&\tim{if }\alpha+2\beta<1,\\
\sigma_{\rm tot}\mu_{\max}\sqrt{1+\log(k+1)}
&\tim{if }\alpha+2\beta=1,\\*[1.5ex]
\sigma_{\rm tot}\mu_{\max}\zeta(\alpha+2\beta)
&\tim{if }\alpha+2\beta>1.\\
\end{array}\right.
\eeqn
In order to obtain \req{detailed-rates-mg} and given \req{nuk-vals} and
\req{thetak-vals}, we now verify which term in the maximum of
\req{Theta-mg} dominates for the different combinations of $\alpha$
and $\alpha+2\beta$.\\
$\bullet$ If $\alpha+2\beta < 1$, then $\alpha<1$
and $\half(1-\alpha)-\beta >0$, and thus
\[
\max\left[(k+1)^{\half(1-\alpha)-\beta},
\sqrt{\log\left((k+1)^{\half(1-\alpha)}\right)}\right]
= (k+1)^{\half(1-\alpha)-\beta},
\]
yielding $\Theta_k = \calO((k+1)^{1-\alpha-\beta})$.\\
$\bullet$ If $\alpha < 1$ and $\alpha+2\beta = 1$,
\[
\max\left[\sqrt{\log(k+1)},\sqrt{\log\left((k+1)^{\half(1-\alpha)}\right)}\right]
= \sqrt{\log(k+1)},
\]
giving $\Theta_k = \calO(\sqrt{\log(k+1)}(k+1)^{\half(1-\alpha})$.\\
$\bullet$ If $\alpha < 1$ and $\alpha+2\beta > 1$,
\[
\max\left[\zeta(\alpha+2\beta),\sqrt{\log\left((k+1)^{\half(1-\alpha)}\right)}\right]
= \sqrt{\log\left((k+1)^{\half(1-\alpha)}\right)}
= \sqrt{\half(1-\alpha)\log(k+1)},
\]
and thus $\Theta_k =\calO(\sqrt{\log(k+1)}(k+1)^{\half(1-\alpha)})$.\\
$\bullet$ If $\alpha = 1$ and $\alpha+2\beta = 1$,
\[
\max\left[\sqrt{\log(k+1)},\sqrt{\log\left(\sqrt{\log(k+1)}\right)}\right]
=\sqrt{\log(k+1)},
\]
implying $\Theta_k =\calO(\log(k+1))$.\\
$\bullet$ If $\alpha = 1$ and $\alpha+2\beta > 1$,
\[
\max\left[\zeta(\alpha+2\beta),\sqrt{\log\left(\sqrt{\log(k+1)}\right)}\right]
=\sqrt{\log\left(\sqrt{\log(k+1)}\right)}
=\sqrt{\half \log(\log(k+1))},
\]
and $\Theta_k =\calO( \sqrt{\log(k+1)}\sqrt{\log(\log(k+1))} )$.\\
$\bullet$ If $\alpha > 1$ and $\alpha+2\beta > 1$,
\[
\max\left[\zeta(\alpha+2\beta),\sqrt{\log(\zeta(\alpha))}\right]
=\calO(1),
\]
and $\Theta_k= \calO(1)$.\\
Inserting these bounds on $\Theta_k$ in \req{thetruerate-alt} then
gives \req{detailed-rates-mg}.

We conclude by noting that \req{ineq1-mom} and \req{ineq2-mom} are
satisfied for all methods considered because the results from
Section~\ref{sec:appl} still hold when the approximate gradient
$\tG_k$ is set to the momentum $M_k$. Moreover \req{ass:sub-add} can
easily be verified for all methods, as replacing $\tG_{k,\ell}$ by a
general matrix $U$ of suitable size in the derivations involving
$\calU_{k,\ell}(\cdot)$ implies that \req{ass:sub-add} holds with
$\kappa_\Box=2$ and $\kappa_\diamond=1$ for all these methods, except that
$\kappa_\diamond = 1/N$ for AdaNorm.

\end{document}